\documentclass{article}

\usepackage{microtype}
\usepackage{graphicx}
\usepackage{booktabs} 
\usepackage{lipsum}

\usepackage{subcaption}
\usepackage{mwe}

\usepackage[utf8]{inputenc}
\usepackage[english]{babel}
\usepackage[dvipsnames]{xcolor}

\usepackage{hyperref}
\usepackage{bbold}
\usepackage{eqnarray,amsmath}
\usepackage{comment}
\usepackage{bbding}

\newcommand{\defeq}{\overset{\text{\tiny def}}{=}}

\usepackage[accepted]{icml2020}

\icmltitlerunning{Tackling Dynamics in Federated Incremental Learning with Variational Embedding Rehearsal}

\begin{document}

\twocolumn[
\icmltitle{Tackling Dynamics in Federated Incremental Learning with Variational Embedding Rehearsal}


\icmlsetsymbol{equal}{*}

\begin{icmlauthorlist}
    \icmlauthor{Taejin Park}{to}
    \icmlauthor{Kenichi Kumatani}{ed}
    \icmlauthor{Dimitrios Dimitriadis}{goo}
\end{icmlauthorlist}

\icmlaffiliation{to}{NVIDIA, Santa Clara, CA, USA}
\icmlaffiliation{ed}{Azure AI, Microsoft, Bellevue, WA, USA}
\icmlaffiliation{goo}{Microsoft Research, Redmond, WA, USA}

\icmlcorrespondingauthor{Dimitrios Dimitriadis}{didimit@microsoft.com}

\icmlkeywords{Incremental Learning, Federated Learning, Variational Embeddings, Rehearsal}

\vskip 0.3in
]



\printAffiliationsAndNotice{}  

\begin{abstract}
Federated Learning is a fast growing area of ML where the training datasets are extremely distributed, all while dynamically changing over time. Models need to be trained on clients' devices without any guarantees for either homogeneity or stationarity of the local private data. The need for continual training has also risen, due to the ever-increasing production of in-task data. However, pursuing both directions at the same time is challenging, since client data privacy is a major constraint, especially for rehearsal methods. Herein, we propose a novel algorithm to address the incremental learning process in an  FL scenario, based on realistic client enrollment scenarios where clients can drop in or out dynamically. We first propose using deep Variational Embeddings that secure the privacy of the client data. Second, we propose a server-side training method that enables a model to rehearse the previously learnt knowledge. Finally, we investigate the performance of federated incremental learning in dynamic client enrollment scenarios. The proposed method shows parity with offline training on domain-incremental learning, addressing challenges in both the dynamic enrollment of clients and the domain shifting of client data.
\end{abstract}

\section{Introduction}
\label{sec:intro}

An ever-increasing need for more training data has created a series of challenges. First, the ML providers need to retain massive amounts of data, all while the models ``consume'' new/fresh data to keep improving. As such, the traditional approach of centralizing all the available data and training a new model every time there is an update is not sustainable. The second challenge is about the clients' data privacy requirements and how this private data can be incorporated in the training set.
A solution to these challenges is to federate the training process, ensuring that model performance doesn't degrade when new data is introduced into the processing pipeline. 
Federated Learning (FL) has gained a lot of attention due to the distributed nature of the algorithms and the flexibility to learn on non-i.i.d data distributions, while preserving a level of privacy~\cite{Li+20, Kairouz+21, wang+21}.
The FL paradigm can address the privacy concerns (especially when models are trained with Differential Privacy)~\cite{DMNS06}. The general principle is based on training different versions of the model on the local data samples, and exchanging only the updates of the model parameters, such as the network parameters or the corresponding gradients. 
\begin{figure*}[htb]
    \centering
    \includegraphics[width=0.8\textwidth]{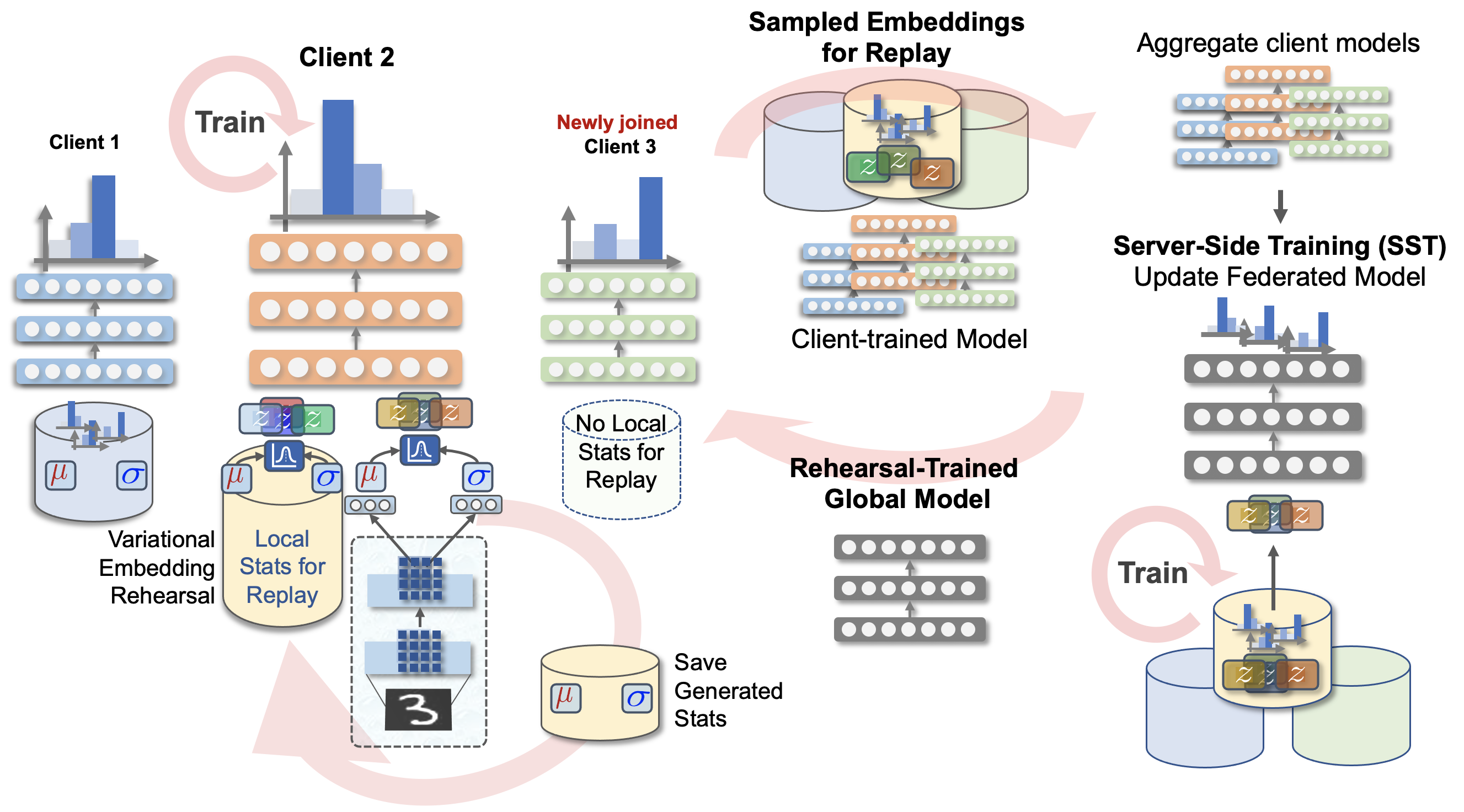}
    \caption{Data flow diagram in the proposed Incremental Learning system in federated learning framework. The Variational Embedding Encoder (VEE) is pretrained on the data from the $1^{st}$ task. The VEE  produces latent Variational embeddings from the local client data, used as input to the classifier. Clients create new  embeddings, sample and transmit them back to the server for rehearsal during SST.}
    \label{fig:block_diagram}
\end{figure*}

Algorithms for FL are designed for model training featuring data parallelism across a large number of nodes, efficiently handling data imbalance, and data sparseness for local training examples,~\cite{KMR15}. It is possible that the data found in each of the clients are coming from vastly different, time-varying data distributions, i.e. these training datasets cannot be considered either i.i.d. or static in time. The possibility of Catastrophic Forgetting becomes substantial due to this dynamic nature of the data influx and the increased diversity between the contributing clients.

Catastrophic Forgetting is the situation when an ML model loses the ability of prediction on  previously trained tasks after training on new data. Catastrophic Forgetting has been considered a critical challenge for neural network based models especially  when these models cannot access previously seen training samples. In general, such learning processes are also called as ``Continual Learning'' (CL),~\cite{Delange+21}, since models are trained sequentially on a stream of unseen data, without access to the past tasks \cite{french1999catastrophic}. CL approaches alleviate the need to store, shuffle and retrain on already seen data. 

There are different scenarios proposed as part of the CL~\cite{hsu2018re}: most notably, the ``\textit{Incremental Domain Learning}'' (IDL), ``\textit{Incremental Class learning''} (ICL) and ``\textit{Incremental Task Learning}'' (ITL). The majority of CL-related algorithms are focused on ITL, where the  model has multiple heads for different tasks. In such a scenario, task-invariant representations of the inputs are expected, and the model is aware of the ``task-specific head''\footnote{The last layer of a classifier is also called  ``classification head''.}. Herein, we focus on the IDL scenario, where the  model does not need to change the classification head, i.e. the  number of output classes remains fixed. Under IDL, the domain of training data changes over time while the model does not have access to the training datasets of the previous tasks. 

Although, research on ``Incremental Learning'' (IL) (or Continual Learning more generally) is rarely considered as part of the Federated Learning framework, there is some related work found in \cite{yoon2020federated}, focused on the ITL scenario. A different approach is learning correlations that are fairly invariant across training tasks, but can hold in unseen testing distributions, e.g., in ``Invariant Risk Minimization''~\cite{ABGL20}, where causal correlations in the training data are learnt. An extension of this work in the FL setup can be found in~\cite{FrTeRi21}.

Herein, we propose a novel approach to the challenge of IL as part of the FL framework. The contributions of this paper are three-fold: We investigate the use of Deep Variational Embeddings,~\cite{JIANG+17}, as part of the IL tasks, more specifically in the IDL scenario. The proposed approach combines the \textit{Embedding Rehearsal} method,~\cite{PSU21}, with the \textit{Variational Embeddings} framework and applied on an IDL scenario. Then, this approach is extended to the Federated Learning setup, where  each of the participating clients communicate task-specific embeddings back to the server. The proposed method adds noise to the representations making it impossible to reverse-engineer the original training samples, all while keeping the performance across tasks consistent. Additionally, we propose ``\textit{Server Side Training}'' (SST), which is novel in the Federated Learning literature. By employing SST,  the representations already collected from clients (and thus retaining that knowledge from the customer data) can be replayed on the server. As such, we no longer need access to the client data for rehearsing. Further, the proposed SST approach allows for future model refinements and/or different model architectures without the need to revisit these previously seen datasets. Finally, this is the first time to our knowledge that realistic node/end-user enrollments scenarios are introduced as part of a Federated Learning system. Herein, we investigate the performance of the proposed Federated Incremental Learning system based on the end-user behavior, opting in and out of the training process.

The similarities and differences with other methods proposed in~\cite{Casado+21, FrTeRi21, hsu2018re} should be highlighted. Based on the latter work, the advantages of the Rehearsal scheme for the IDL tasks have become apparent. One of the Rehearsal variations is based on deep embeddings\footnote{This scheme is not realistic for FL scenarios due to privacy constraints.}. Herein, the proposed algorithm is based on the ``\textit{Variational Embedding Encoder}'' (VEE) trained on the classification task (rather than minimizing a reconstruction error as in other literature). After training, the encoder remains fixed and it is used as the front-end for the classifier across different tasks. Although the encoder is optimized on held-out data,  the embeddings show a relative task-invariance (especially when combined with SST). The use of Variational Embeddings allows the decoupling of the input features and the output embeddings, and as such privacy is ensured.
On the other hand, the embedding extractor in~\cite{FrTeRi21} is always updated during training. Finally, the SST step is introduced as part of the federated learning task, where rehearsal takes place. Finally, the proposed method in~\cite{Casado+21} follows a different direction of revisiting the FedAvg algorithm to incorporate different information at particular iterations.
The rest of the paper is organized as follows. Section~\ref{sec:background} reviews basic techniques that we extend for federated IL (FIL). In Section~\ref{sec:proposed}, we describe our proposed FIL method to avoid catastrophic forgetting without privacy violation. In Section~\ref{sec:dynamic_enroll}, we newly introduce the dynamic enrollment scenarios to analyze FIL performance. Section~\ref{sec:experiments} describes experiment results in the conventional IL setting as well as the dynamic enrollment scenario. We conclude this work in Section~\ref{sec:conclusions}. 

\section{Background}
\label{sec:background}

\subsection{Federated Averaging -- FedAvg}
\label{subsec:fedAvg}

A simple algorithm like the \emph{Federated Averaging} (FedAvg) algorithm,~\cite{KMR15, McMahan+17} is the first and perhaps the most widely used FL training algorithm. The server~\footnote{The server is the orchestrator, i.e., sampling clients, aggregating and updating models, etc} samples $\mathcal{M}_T\subset \mathcal{N}$  of the available $\mathcal{N}$ devices and sends the model $\mathbf{w}^{(s)}_T$ at that current iteration $T$. Each client $j$, with $j\in \mathcal{M}_T$ has a version of the model $\mathbf{w}^{(j)}_T$ where it is locally updated with the segregated local data. The size of the available data $\mathcal{D}^{(j)}_T$,  per iteration $T$ and client $j$, is expected to differ and as such, $N^{(j)}_T=|\mathcal{D}^{(j)}_T|$ is the size of available local training samples. After running $\mathcal{E}$ steps of SGD, the updated models $\hat{\mathbf{w}}^{(j)}_T$ are sent back to the server. The new model $\mathbf{w}_{T+1}$ is given by
\begin{equation}
    \mathbf{w}_{T+1} \leftarrow \frac{1}{\sum_{j\in\mathcal{M}_T} {N^{(j)}_T}}\sum_{j\in\mathcal{M}_T}{N^{(j)}_T \hat{\mathbf{w}}^{(j)}_T}
\label{eq:fedavg}
\end{equation}
The global model in iteration~$T+1$ is the weighted average of the locally updated models of the previous iteration~$T$.

Although, FedAvg has certain drawbacks, it is the training algorithm of choice throughout this  proposed work. The focus of our paper is less on the federation algorithms and mostly on mitigating the problem of Catastrophic Forgetting in the context of FL. As such, the golden standard of FL training, i.e. the FedAvg, is used for our experiments.

\subsection{Variational Autoencoders (VAEs)}
\label{subsec:VAE}

\begin{figure}[!t]
    \centering
    \includegraphics[width=0.5\textwidth]{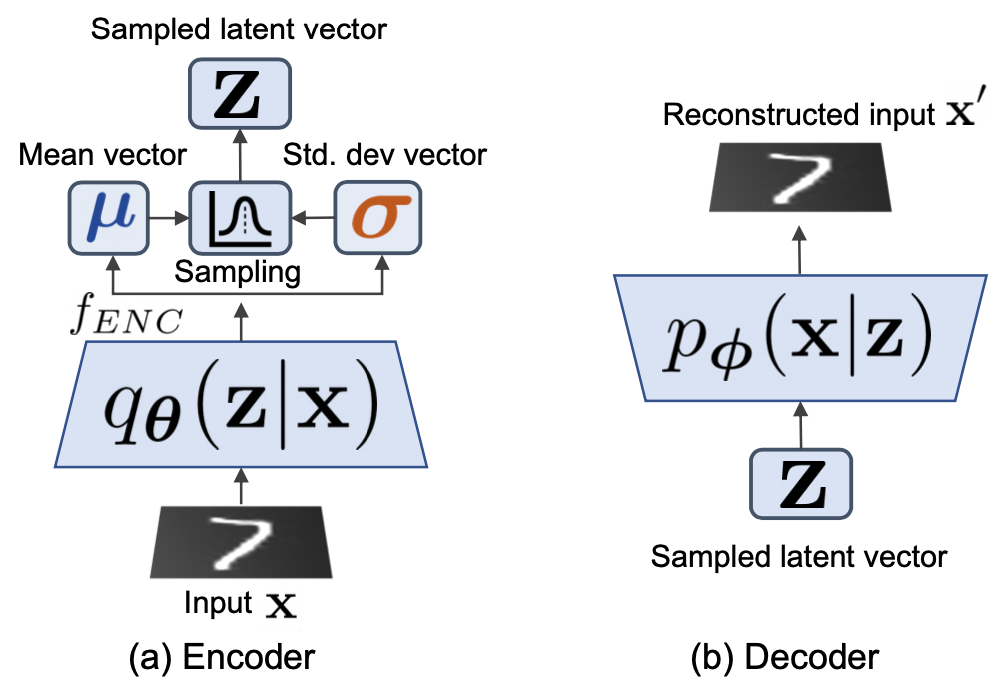}
    \caption{Variational Autoencoder Architecture: (a) the Encoder estimates latent variables $\mathbf{z}$ from the input $\mathbf{x}$ and (b) the Decoder reconstructs the input $\mathbf{x}$ variables from $\mathbf{z}$.}
    \label{fig:VAE_enc_dec}
    \vspace{-2ex}
\end{figure}

\textit{Variational Autoencoders} (VAEs) encode inputs as trainable distributions instead of points in a latent space~\cite{kingma2013auto, rezende2014stochastic}. The latent space  is typically regularized by constraining the parameters for the distributions generated by the encoder to match a Gaussian distribution. 
As shown in Fig.~\ref{fig:VAE_enc_dec}(a), the VAE encodes the input data~$\mathbf{x} \in \mathbb{R}^n$ with~$q_{\boldsymbol{\theta}}(\mathbf{z}|\mathbf{x})$ into the latent variable $\mathbf{z} \in\mathbb{R}^d$ (usually $d\ll n$) and reconstructs the data~$\mathbf{\acute{x}}$ from the latent variable $\mathbf{z}$ through the decoder $p_{\boldsymbol{\phi}}(\mathbf{x}|\mathbf{z})$, in Fig.~\ref{fig:VAE_enc_dec}(b). 
The latent variable $\mathbf{z}$ is sampled from a learnt prior distribution $p(\mathbf{z})$ and the reconstructed output~$\mathbf{\acute{x}}$ is sampled from the conditional likelihood distribution $p_{\boldsymbol{\phi}}(\mathbf{x}|\mathbf{z})$. 
Based on this model\footnote{Contrary to the common autoencoder approach where the encoder/decoder are deterministic.}, the \emph{probabilistic decoder} is defined by $p_{\boldsymbol{\phi}}(\mathbf{x}|\mathbf{z})$ describing the distribution of the decoded variable given the encoded one.
Similarly, the \emph{probabilistic encoder} is defined by $q_{\boldsymbol{\theta}}(\mathbf{z}|\mathbf{x})$, describing the distribution of the encoded variable given the decoded one. 
The regularization of the latent space is defined by the encoded representations $\mathbf{z}$ in the latent space following the prior distribution $p(\mathbf{z})$. The generative process of VAEs is described with the following terms: marginal likelihood $p_{\boldsymbol{\phi}}(\mathbf{x})$, complex likelihood $p_{\boldsymbol{\phi}}(\mathbf{x}|\mathbf{z})$ and posterior $p_{\boldsymbol{\phi}}(\mathbf{z}|\mathbf{x})$.

Since calculating the marginal likelihood~$p_{\boldsymbol{\phi}}(\mathbf{x})$ is often computationally intractable, the VAE framework approximates the distribution $p_{\boldsymbol{\phi}}(\mathbf{z}|\mathbf{x})$ by introducing the function $q_{\boldsymbol{\theta}}(\mathbf{z}|\mathbf{x})$, where $\boldsymbol{\theta}$ is defined as the set of parameters in the encoder. Thus, by introducing the approximation, $p_{\boldsymbol{\phi}}(\mathbf{z}|\mathbf{x}) \approx q_{\boldsymbol{\theta}}(\mathbf{z}|\mathbf{x})$, the overall problem can be modeled within the autoencoder framework, where the conditional likelihood $p_{\boldsymbol{\phi}}(\mathbf{x}|\mathbf{z})$ is modeled by the decoder part and the posterior distribution $q_{\boldsymbol{\theta}}(z|x)$ is modeled by the encoder part. Accordingly, VAE is trained by jointly minimizing the generative model parameters $\boldsymbol{\phi}$ to reduce the reconstruction error between the inputs and outputs of the network, and $\boldsymbol{\theta}$ to match $q_{\boldsymbol{\theta}}(\mathbf{z}|\mathbf{x})$ as close as possible  to $p_{\boldsymbol{\phi}}(\mathbf{z}|\mathbf{x})$. The goal now becomes the maximization of the marginal likelihood of  $p_{\boldsymbol{\phi}}(\mathbf{x})$.  
Thus, the loss function for training the autoencoder is derived from the log marginal likelihood of $p_{\boldsymbol{\phi}}(\mathbf{x})$,~\cite{kingma2019introduction},
\begin{equation}
    \begin{split}
    \label{eq:deriv_ELBO}
 \log p_{\boldsymbol{\phi}}(\mathbf{x}) 
 =&\mathbb{E}_{\mathbf{z}}\left[\log p_{\boldsymbol{\phi}}\left(\mathbf{x}|\mathbf{z}\right)\right] \\ 
 &-D_{K L}\left(q_{\boldsymbol{\theta}}\left(\mathbf{z}|\mathbf{x}\right) \| p_{\boldsymbol{\phi}}(\mathbf{z})\right)\\ 
 &+D_{K L}\left(q_{\boldsymbol{\theta}}\left(\mathbf{z}|\mathbf{x}\right) \| p_{\boldsymbol{\phi}}\left(\mathbf{z}|\mathbf{x}\right)\right)
\end{split}
\end{equation}
In the above equation, it is not feasible to compute the third term since $p_{\boldsymbol{\phi}}\left(\mathbf{z}|\mathbf{x}\right)$ is intractable. Since $p_{\boldsymbol{\phi}}\left(\mathbf{z}|\mathbf{x}\right)\geq 0$ based on the property of the Kullback Leibler (KL) divergence, 
the VAE loss, i.e. $\max_{\boldsymbol{\phi}} \log p_{\boldsymbol{\phi}}(\mathbf{x})$, is optimized, 
by maximimizing the first two terms. These two terms are usually referred to as the ``Evidence Lower Bound'' (ELBO) or Variational lower bound,
\begin{equation}
\label{eq:ELBO}
\text{ELBO} = \mathbb{E}_{\mathbf{z}}\left[\log p_{\boldsymbol{\phi}}\left(\mathbf{x}|\mathbf{z}\right)\right] \\ 
 -D_{K L}\left(q_{\boldsymbol{\theta}}\left(\mathbf{z}| \mathbf{x}\right) \| p_{\boldsymbol{\phi}}(\mathbf{z})\right)\\ 
\end{equation}

Therefore, the training of VAE is by maximizing ELBO  in Eq.~(\ref{eq:ELBO}) and the maximization of ELBO value can lead to the two following things: First, it will approximately maximize $p_{\boldsymbol{\phi}}(\mathbf{x})$ so that the generated samples  become closer to the actual training samples. Second, maximizing ELBO will minimize the KL divergence of $p_{\boldsymbol{\phi}}(\mathbf{z}|\mathbf{x})$ and $q_{\boldsymbol{\theta}}(\mathbf{z}|\mathbf{x})$, so the $q_{\boldsymbol{\theta}}(\mathbf{z}|\mathbf{x})$ gets closer to the distribution $p_{\boldsymbol{\phi}}(\mathbf{z}|\mathbf{x})$.

An interesting part of the VAEs is the \textit{reparameterization trick}. This reparameterization trick leads to a differentiable generative model (encoder) with respect to both $\boldsymbol{\theta}$ and $\boldsymbol{\phi}$ through a change of variables. A common choice of distribution for the reparameterization trick is the Normal distribution since it leads to an analytic form of KL-divergence. Thus, by employing the Normal distribution, the change of variables can be described as below:
\begin{equation}
\label{eq:VAE_sampling}
\begin{aligned}
\boldsymbol{\epsilon} & \sim \mathcal{N}(0, \mathbf{I}) \\
(\boldsymbol{\mu}, \log \boldsymbol{\sigma}) &\leftarrow f_{ENC}(\mathbf{x}; \boldsymbol{\theta}) \\
\mathbf{z} &=\boldsymbol{\mu}+\boldsymbol{\sigma} \odot \boldsymbol{\epsilon}
\end{aligned}
\end{equation}
where $\odot$ is the element-wise product and $f_{ENC}(\ \cdot\ ;\boldsymbol{\theta})$ is the encoder. Thus, if we constraint the distribution of the sampled latent vector $p_{\boldsymbol{\phi}}(z)$ to follow the unit Gaussian, the loss function we optimize in order to train VAE becomes:
\begin{equation}
\begin{split}
\label{eq:VAE_loss}
\mathcal{L}_{\boldsymbol{\theta}, \boldsymbol{\phi}}(\mathbf{x})=
&-\mathbb{E}_{\mathbf{z}}\left[\log p_{\boldsymbol{\phi}}\left(\mathbf{x} | \mathbf{z}\right)\right] \\ 
&+D_{K L}\left(q_{\boldsymbol{\theta}}\left(\mathbf{z} | \mathbf{x}\right) \| \mathcal{N}(0, \mathbf{I})
 \right)\\ 
 \end{split}
\end{equation}
Finally, if we assume that $\mathbf{x}$ follows normal distribution, we can arrange the first term (reconstruction loss) in mean-square error (MSE) form. In addition, if we replace the encoder output $q_{\boldsymbol{\theta}}(\mathbf{z} | \mathbf{x})$ with the distribution we described in Eq.~(\ref{eq:VAE_sampling}), the Eq.~(\ref{eq:VAE_MSE_loss}) can be rewritten as follows:
\begin{equation}
\begin{split}
\label{eq:VAE_MSE_loss}
\mathcal{L}_{\boldsymbol{\theta}, \boldsymbol{\phi}}(\mathbf{x})=
&\mathbb{E}_{\mathbf{\mathbf{z}}}\left[
\left(\frac{\|\mathbf{x}-f_{DEC}(\mathbf{z};\boldsymbol{\phi})\|^{2}}{2 c}\right)
\right] \\ 
&+D_{K L}\left( \mathcal{N}(\boldsymbol{\mu}, \boldsymbol{\sigma}) \| \mathcal{N}(0, \mathbf{I})
\right),
\end{split}
\end{equation}
where $c$ is the variance of the likelihood function and $f_{DEC}(\ \cdot \ ;\boldsymbol{\phi})$ is the decoder, and $c$ compensates for the reconstruction loss and the KL-divergence.




\subsection{Embedding Based Rehearsal (EBR)}
\label{subsec:embeddings}

The approach of generating or extracting representations from input features as a  mitigation strategy for Catastrophic Forgetting has been well studied,~\cite{JaWh19, FrTeRi21, SLKK17, RKSL17}. The main idea is to either extract task invariant representations or generate new embeddings and use them for Naive Rehearsal, an idea called \emph{Embedding Based Rehearsal} (EBR). Several approaches have been proposed for training the embedding extractor $f_{EBR}(\cdot;\boldsymbol{\theta})$ under different criteria and expected properties. Since it is  almost impossible to cover all of these different approaches, we herein examine the basic EBR approach. Despite its simplicity, experimental results validate its efficiency, as shown in Section~\ref{subsec:exp_ebr}.

The goal is to train the encoder $\boldsymbol{\theta}$ of the embedding extractor $f_{EBR}(\mathbf{x}; \boldsymbol{\theta}): \mathbb{R}^n\rightarrow \mathbb{R}^d$, where $(\mathbf{x}, y)$ with $\mathbf{x}\subset \mathcal{X}: \mathbb{R}^n$ are the input features drawn from a dataset $\mathcal{D}:(\mathcal{X}, \mathcal{Y})$,  labels $y\subset\mathcal{Y}:\mathbb{R}$, and $\mathbf{z}_{EBR}\subset \mathbb{R}^d$ the output latent embeddings, i.e. $\mathbf{z}_{EBR}=f_{EBR}(\mathbf{x}; \boldsymbol{\theta}),\ \forall \mathbf{x}\in\mathcal{X}$. 
The function $f_{EBR}(\cdot; \boldsymbol{\theta})$ is cascaded to the classifier $f_{CLS}(\ \cdot\ ; \mathbf{w}): \mathbb{R^d} \rightarrow \mathcal{Y}$, with $\mathbf{w}$ the classifier parameters, 
and trained by back-propagating the classifier empirical loss $\mathcal{L}_{CE}(\cdot)$\footnote{Cross-Entropy loss is presented here without loss of generality.},
\begin{equation}
    \mathcal{L}_{CE}\left(\mathbf{x},y\right) =-y\log(f_{CLS}\left(f_{EBR}(\mathbf{x};\boldsymbol{\theta}); \mathbf{w}\right),
\label{eq:loss}
\end{equation}
with $\mathbf{x}\in \mathcal{D}_h$. The extractor $f_{EBR}(\cdot; \boldsymbol{\theta})$ 
can be trained with either in-task  or  held-out data $\mathcal{D}_h \subset \mathcal{D}$ before fixing the weights of $\boldsymbol{\theta}^*$\footnote{Encoder parameters $\boldsymbol{\theta}^*$ are optimized.}. 
The $f_{EBR}^*(\cdot)\defeq f_{EBR}(\ \cdot\ ;\boldsymbol{\theta}^*)$ is then used for all the subsequent tasks without updating it. The features/embeddings used for training local models are now extracted by the extractor. These embeddings are then sampled and replayed as part of the Naive Rehearsal strategy. 

In the case of FL, there is an additional step where the FL server collects the locally updated models and embeddings from the clients for rehearsal. Transmitting the deterministic embeddings $f_{EBR}^*(\cdot)$ may violate client's privacy since such an embedding has a one-to-one correspondence between the input signal and output embedding and can be reverse engineered to identify the raw data provided there is access to the encoder, e.g. in the case of face identification \cite{zhmoginov2016inverting, cole2017synthesizing}. 



\section{Variational Embedding Rehearsal and Server-Side Training}
\label{sec:proposed}

\subsection{Variational Embedding Rehearsal (VER)}
\label{subsec:MGN}

As stated in the previous section, although EBR is an effective strategy for IL scenarios, it is not tractable for FL applications due to privacy concerns. Therefore, we need to provide  privacy guarantees for the input samples before employing an embedding-based rehearsal method for FL.


To tackle this problem, we propose adding noise to the embeddings, but without affecting the classification performance. The added noise addresses the drawbacks of EBR by creating private representations. To do that, we replace the decoder part $f_{DEC}(\ \cdot \ ;\boldsymbol{\phi})$ of VAEs with a classifier  $f_{CLS}(\ \cdot\ ;\mathbf{w}_T)$, where $\mathbf{w}_T$ the classifier parameters in iteration $T$ (similar to Section~\ref{subsec:embeddings}). The proposed scheme is based on the ``Variational Embeddings'' and the  overall proposed system is called ``\textit{Variational Embedding Rehearsal}'' (VER). Accordingly, we refer to the encoder part of VER system as ``\textit{Variational Embedding Encoder}'' (VEE) and henceforth, denoted as $f^*_{VEE}(\cdot)$\footnote{As before, $f^*_{VEE}(\cdot)\defeq f_{VEE}(\ \cdot\ ; \boldsymbol{\theta}^*)$, where $\boldsymbol{\theta}^*$ are the trained encoder parameters.}. The difference between EBR and VER lies in the encoder used.

In terms of the encoder parameters $\boldsymbol{\theta}$, the $f_{VEE}(\ \cdot\ ;\boldsymbol{\theta})$ is identical to the encoder part in VAEs. However, contrary to the VAE case, $f_{VEE}(\ \cdot\ ;\boldsymbol{\theta})$ learns the distribution generating $\mathbf{z}$ that minimizes the classification error (herein, Cross-Entropy), while constraining the distribution of $\mathbf{z}$ to the prior $p(\mathbf{z})$. Thus, $f_{VEE}^*(\cdot)$ generates stochastic representations, preserving the classification performance without any privacy violation, contrary to the deterministic embedding produced by $f_{EBR}^*(\cdot)$.




%
Thus, the generated Variational embeddings $\mathbf{z}$ cannot be deterministically mapped to the input data and they can still be used to rehearse on the past samples. Following Section~\ref{subsec:VAE}, the Variational embeddings $\mathbf{z}_{T}$ that is generated from $f^*_{VEE}(\cdot)$ is:
\begin{equation}
\label{eq:function_VEE}
\mathbf{z}_{T} = f^*_{VEE}(\mathbf{x}_{T}),
\end{equation}
where $f^*_{VEE}(\cdot)$ includes the following operations described in Eq.(\ref{eq:VAE_sampling}).
The latent variables $\mathbf{z}_T$ are associated with the labels $y_T$ from the $\mathcal{D}_T$ and saved for training the classifier $f_{CLS}(\ \cdot\ ;\mathbf{w}_T)$ replaying in future iterations and finally sent to the server for SST, as described in Section~\ref{subsec:SST}.

The VER partially replicates the structure of VAE but the decoder part is replaced by an in-task classifier -- the VER system is embedded in the classification task rather than the decoding task. Since, we are not performing any decoding (but the encoding part remains) the loss function for the VER is revisited by switching the reconstruction loss with a cross-entropy loss for the classification task. As such, the loss function for VER is:
\begin{equation}
\begin{split}
\label{eq:VER_loss}
\mathcal{L}_{\boldsymbol{\theta}, \boldsymbol{\phi}}(\mathbf{x}_T) &= 
\mathbb{E}_{\mathbf{z}_{T}}\left[
\mathcal{L}_{CE}(f_{CLS}\left(\mathbf{z}_T;\mathbf{w}_T\right), y_T)
\right]\\
&+ \beta D_{K L}\left(
\mathcal{N}(\boldsymbol{\mu}_T, \boldsymbol{\sigma}_T)  \| \mathcal{N}(0, \mathbf{I})
\right),
\end{split}
\end{equation}
where $\mathcal{L}_{CE}$ from Eq.~\ref{eq:loss}, $y_{T}$ is the ground-truth label of the input sample $\mathbf{x}_T$ and $\beta$ is a coefficient that determines the ratio between the two terms in Eq.~(\ref{eq:VER_loss}).





\subsection{Federating Incremental Learning with VER}
\label{subsec:fedMGN}

The proposed ``\textit{Federated Incremental Learning with Variational Embeddings Rehearshal}'' (FIL-VER) is an extension of the VER method in the FL setting. 
The high-level structure of the proposed client- and server-side components is described in Figure \ref{fig:block_diagram}. 
The goal of our FIL-VER is rehearsing a subset of the stochastic representations produced with the VEE on the client side instead of relaying the raw data such as the ``Naive Rehearsal'' approach~\cite{hsu2018re}. 

The pretrained encoder $f^*_{VEE}(\cdot)$ is shared across all clients $j$, where $j\in \mathcal{M}_T$, contains the local data $\mathcal{D}^{(j)}_T=(\mathbf{x}^{(j)}_T, y^{(j)}_T)$ with $\mathcal{D}^{(j)}_T: \mathbb{R}^n \times \mathbb{R}$ 
and $T$ is the current FL training iteration. We assume that the local data $\mathcal{D}^{(j)}_T$ can change in every iteration $T$ and their distribution remains non-i.i.d., as described in Section~\ref{subsec:fedAvg}.

Versions of the current model $\mathbf{w}_T$ are trained on a number of clients $j$, according to FedAvg. These models are trained on the output of the $f_{VEE}^*(\cdot)$ encoder, while estimating the embedding statistics $\left(\mu^{(j)}_T, (\sigma_T^{(j)})^2\right)$. Then, the models are transmitted back to the server for the model aggregation step, Eq.~\ref{eq:fedavg}. In parallel, the clients sample the local data, to create the corresponding embeddings with $f^*_{VEE}(\cdot)$ and  transmit them to the server, i.e. $\mathcal{R}_{T,Reh.}^{(j)}: \mathbb{R}^d \times \mathbb{R}$, consisting of the pairs  $(\mathbf{z}^{(j)}_T, y_T^{(j)})$, with $\mathbf{z}^{(j)}_T$ the sampled latent embeddings and $y^{(j)}_T$ the corresponding labels. Now, the server has the updated models and a set $\mathcal{R}_{T,Reh.}$ of latent embeddings it can use for rehearsal,
\begin{equation}
    \mathcal{R}_{T,Reh.}  = \bigcup\limits_{j\in \mathcal{M}_T} \mathcal{R}^{(j)}_{T, Reh.}
\end{equation}

The clients store the locally estimated statistics~$\mu^{(j)}_t$ and~$\sigma^{(j)}_t$ from previous training iterations $t\leq T$, and use them to produce a rehearsal set $\mathcal{R}^{(j)}_{T,Reh.}$ for training. 
The training set for the $j^{th}$ local model $\mathbf{w}_T^{(j)}$ consists of randomly mixed~$\mathcal{R}^{(j)}_{T,Reh.}$ (based on the previous statistics) and a new set of embeddings $f^*_{VEE}(\mathbf{x}^{(j)}_T)$, where $\mathbf{x}^{(j)}_T \in \mathcal{D}^{(j)} _T$. 


\subsection{Server Side Training}
\label{subsec:SST}

An additional step in the proposed algorithm is the ``\textit{Server Side Training}'' (SST), where we replay data based on the statistics shared by the clients of the previous iterations -- this step is held after aggregating the federated local models.
The clients send back the trained models $\hat{\mathbf{w}}_T^{(j)}$ and $\mathcal{R}_{T,Reh.}^{(j)}$, as described in Section~\ref{subsec:fedMGN}. 
After the model aggregation, a rehearsal step, based on both the newly arrived anonymized representations and the saved anonymized representations, is held and the $\mathbf{w}_{s+1}$ is updated, where $s$ is the local rehearsal iteration and $s\in[1,\ S_{max}]$, with $S_{max}$ the max. number of iterations for rehearsal. Finally, the model is distributed to the clients for the next iteration. Herein, the number of rehearsing training examples always remains constant but other strategies can be implemented.

The goal behind \textit{SST}, i.e. pairing FL training with a rehearsal step, is to preserve the privacy of the client data while addressing Catastrophic Forgetting. The model learns from all the local data found on the clients and in parallel, replay some of that on the server. This step reminds the model of previously seen training samples, that are not accessible anymore on the clients. 
As such, SST enables robust system performance against domain shifting of clients' data or sparse enrollment of these clients. The dynamic client enrollment scenarios are examined in Section~\ref{sec:dynamic_enroll}.

\begin{figure}[!ht]
    \centering
    \includegraphics[width=0.5\textwidth]{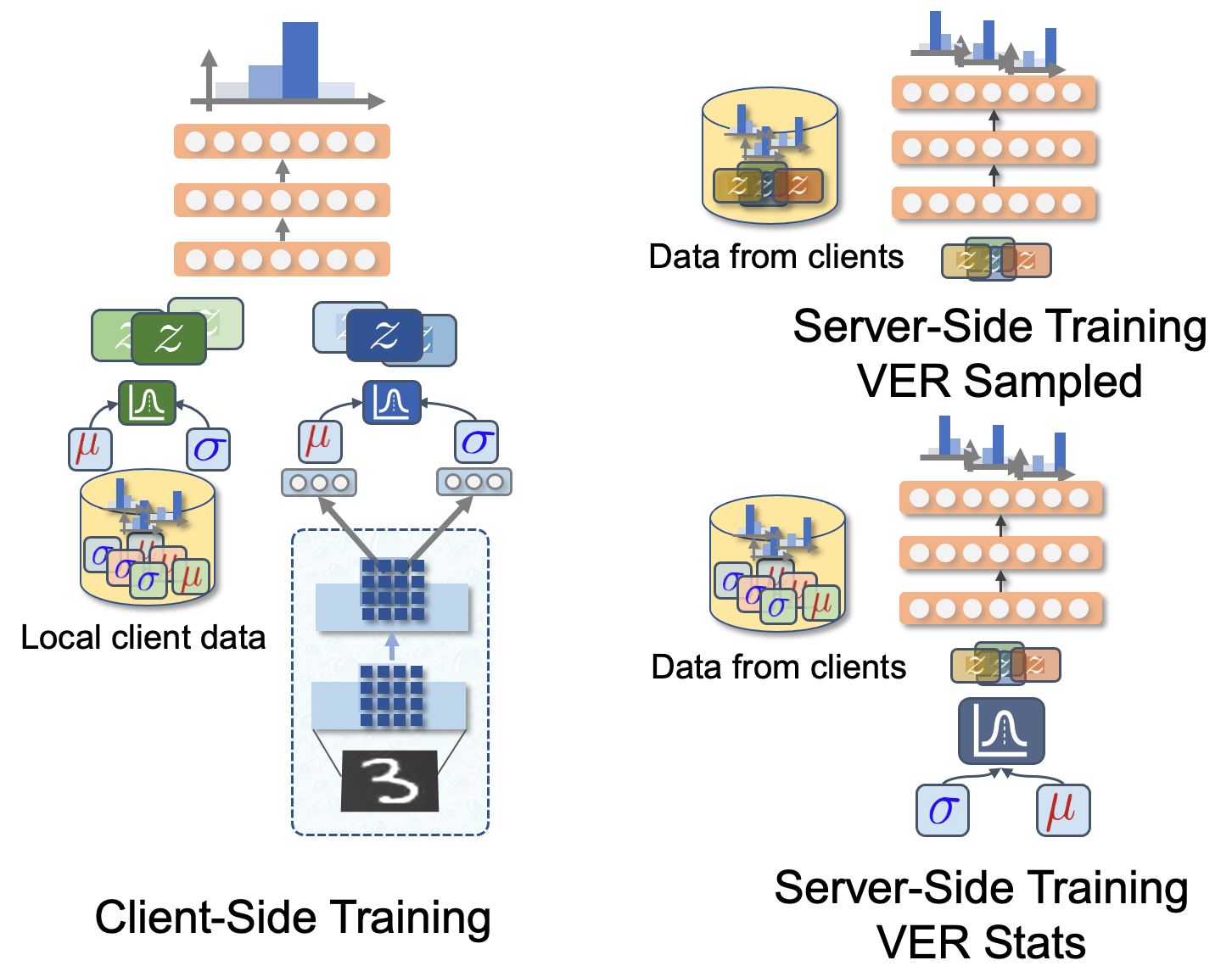}
    \caption{The concept of client-side training and SST. VER-Sampled case is our proposed method where client sends sampled VER vectors. On the other hand, VER-Stats is a framework we created for comparison where clients send to the server mean and variance values, which can infringe local privacy.}
    \vspace{-0.1in}
    \label{fig:CST_SST}
\end{figure}

We investigate two different scenarios for VER:
1) server receiving the \textit{VER Stats}, the encoder statistics, i.e., means and variances and 2) sever collecting embedding estimates called \textit{VER Sampled}.  
Notice that the \textit{VER Stats} method will violate client's privacy since the sample statistics are directly obtained from the clients' encoder without any noise addition. These statistics shared with the server are identity-traceable just like the embeddings in \textit{EBR} method, breaching privacy constraints.
Figure~\ref{fig:CST_SST} illustrates the conceptual flow chart of client-side training and SST with the \textit{VER Stats} and \textit{VER Sampled} methods. 

\begin{algorithm}[!ht]
\caption{Federated Incremental Learning with VER and SST}
\label{alg:MGN_SST}
\begin{algorithmic}[1]
    \item $T_{max}$, max number of FL iterations, $\mathcal{N}$ number of available clients, $\mathcal{M}_T$ sampled clients per iteration, $\mathcal{E}$ local training iterations per client, $S_{max}$ max number of Naive Rehearsal iterations, $\nabla_\mathbf{w} \mathcal{L}(\cdot)$ gradient of local loss function $\mathcal{L}(\cdot)$, $\eta_T$ the learning rate at iteration $T$, and $\eta_s$ the learning rate for SST model updates 
    \vspace{0.25cm}
    \STATE {\bfseries Function \it{FIL-VER} } Input: $w_T^{(s)},\ f^*_{VEE}(\cdot), \mathcal{D}_T$
    \WHILE{Model $\mathbf{w}_T$ has not converged or $T\leq T_{max}$ }
        \STATE Choose $\mathcal{M}_T\subset \mathcal{N}$
        \FOR{$j \in \mathcal{M}_T$}
            \STATE $\mathbf{w}_{T,0}^{(j)}= \mathbf{w}_T^{(s)}$
            \FOR{$i\ in\ [0,\mathcal{E}]$}
                \STATE $\mathbf{w}^{(j)}_{T,i+1} \leftarrow  \mathbf{w}^{(j)}_{T,i}-\eta_T  \nabla_\mathbf{w}\mathcal{L}\left(ENC(\mathbf{x}_T^{(j)})\right)$, \\ where $\forall \mathbf{x}_T^{(j)}\in \mathcal{D}_T^{(j)}$
            \ENDFOR
        \ENDFOR
        \STATE $\mathcal{R}_{T, Reh.} = \bigcup_j \mathcal{R}_{T, Reh.}^{(j)}$, where $\mathcal{R}_{T, Reh.}^{(j)}$ randomly sampled from $\mathcal{R}_T^{(j)}$
        \STATE $\mathbf{w}_{T+1} \leftarrow \frac{1}{\sum_j N^{(j)}_T}\sum_j{N^{(j)}_T \hat{\mathbf{w}}^{(j)}_T}$,\ \ \ as in Eq.~\ref{eq:fedavg}
        \STATE $\mathbf{w}_{T+1}=$ {\bfseries \it{SST}}$(\mathbf{w}_{T+1}, \mathcal{R}_{T,Reh.})$
    \ENDWHILE
\end{algorithmic}
\vspace{0.25cm}
\begin{algorithmic}[1]
    \STATE {\bfseries Function \it{ENC} }(Input: $\mathcal{D}_T$)
    \STATE $ \boldsymbol{\mathbf{z}} \leftarrow f^*_{VEE}(\mathbf{x})$
    ,  where $(\mathbf{x},y)\in \mathcal{D}^{(j)}_T$
    \STATE Return $\mathcal{R}^{(j)}_T,\ \ \  \forall (\mathbf{z},y)\in \mathcal{R}^{(j)}_T$ 
\end{algorithmic}
\vspace{0.25cm}
\begin{algorithmic}[1]
    \STATE {\bfseries Function \it{SST}: }( $\mathbf{w}_T,\ \mathcal{R}_{T, Reh.}$)
    \STATE $\mathbf{w}_0=\mathbf{w}_T$
    \FOR{$s\ in\ [0,S_{max}]$}
        \STATE $\mathbf{w}_{s+1} \leftarrow \mathbf{w}_{s}-\eta_s  \nabla_\mathbf{w}\mathcal{L}\left(\mathbf{z}_T\right)$,\ where $\text{ for }\mathbf{z}_T\in \mathcal{R}_{T, Reh.}$
    \ENDFOR
    \STATE Return $\mathbf{w}_{S_{max}}$
\end{algorithmic}
\end{algorithm}

\section{Dynamic Client Enrollment Scenarios}
\label{sec:dynamic_enroll}

\begin{figure}[htb]
    \centering
    \includegraphics[width=0.47\textwidth]{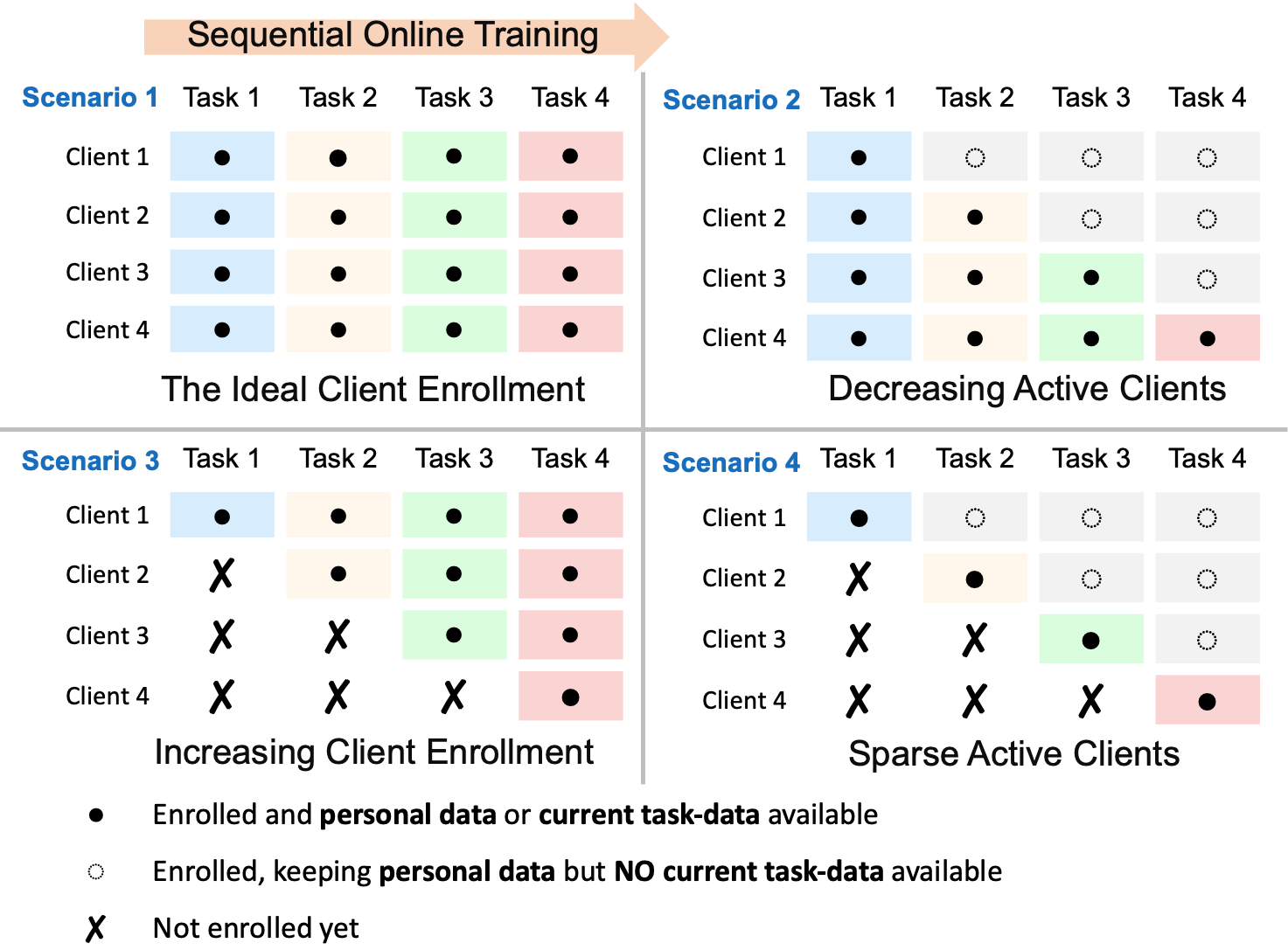}
    \vspace{-1ex}
    \caption{Four types of dynamic client enrollment scenarios.}    
    \vspace{-0.15in}
    \label{fig:dynamic_enrollment}
\end{figure}

A novel experimental setup is herein introduced, where the clients' dynamic registration and participation in a federated learning system is investigated. 
In typical federated learning scenarios, there are two basic assumptions: first, it is expected   all clients remain enrolled/available while the federated learning system is running, and second each of the client data distributions doesn't change over time. Herein, we introduce run-time scenarios, where the clients can register for or withdraw from the service, called the ``\textit{Dynamic Client Enrollment}'' (DCE) scenarios. As such, we have identified four different scenarios depending on the patterns of the client enrollment, based on the three distinct states of each client\footnote{We assume that a client's state doesn't change within tasks.}, i.e., ``\textit{Active}'', ``\textit{No Longer Active}'' and ``\textit{Not Enrolled Yet}'', as  in Fig.~\ref{fig:dynamic_enrollment}. The state marked as a ``$\bullet$'' indicates the ``Active'' state where the client is producing local data and the system can access all of the client's private data. The ``$\bullet$'' means we can take full advantage of the client $j$'s data at task $n$ (and all previously participated tasks). The, ``$\circ$'' indicates clients no longer using the service and as such,  rehearsal data only from their previous tasks can be used. The third state is labeled with ``{\small \XSolidBrush}'' where the client has not yet enrolled  but will enroll later in the training process. In this case, the FL system can access the data only after the client enrolls, which is that case is indicated as ``$\bullet$''. Since the cross mark indicates a non-existent client slot, the client state cannot switch back to ``Not Enrolled yet'' state but only can switch to  ``No Longer Active'', which is indicated by the ``$\circ$''.


\paragraph{Scenario 1: Fully Enrolled}
The ``Fully Enrolled'' scenario is similar to a typical federated learning process, where all clients are active and keep generating new client-side data as training progresses. This scenario is defined as ``all clients are enrolled from the initiation of the training and stay enrolled until the end of training''. In this case, the FL system does not suffer from missing client data unlike the rest of the scenarios. We consider the ``Fully Enrolled'' scenario as the baseline system, where the ability of remembering previous tasks is bench-marked. 

\paragraph{Scenario 2: Decreasing Client base}
This scenario is based on the assumption that some enrolled clients drop out after some iterations. Although some of the clients switch from ``Active'' to ``No Longer Active'',  their data (from previous tasks) are still used for rehearsal. This scenario can be described as ``some clients were previously active but they have opted out of the current training cycle''. Performance on early tasks isn't significantly affected but it will degrade in later tasks when training data appear sparser. 

\paragraph{Scenario 3: Increasing Client base}
This is the case when an increasing number of enrolled clients is participating, 
contrary to the \textit{Scenario 2}. Some of the clients decide to enroll the process after  training has started, and as such, they cannot contribute to the tasks they haven't participated. Since the system has an increasing volume of training data, the model performance is expected to gradually increase for later tasks. On the contrary, performance will lag for early tasks, since the rehearsal SST step cannot benefit from more training examples.

\paragraph{Scenario 4: Scattered Client data}
Finally, \textit{Scenario 4} is the most base learning scenario, where each client opts in for a relatively short time window and as such, the contributing data is sparse. The FL system only acquires data for a particular task, and the amount of training and rehearsal data is decreased. The overall validation accuracy is expected to be relatively lower than any other scenario, due to this sparseness of training examples.

\section{Experiments}
\label{sec:experiments}
The architecture of the classifier $f_{CLS}$ is a 2-layer multilayer perceptron with 1000 hidden units per layer and ReLU activations are used. Both $f_{EBR}$ and $f_{VEE}$ are two layers of convolutional neural networks (CNNs), where each layer has 5$\times$5 kernel size and ReLU activation. Max pooling is performed with a 2$\times$2 window and stride of 2. CNN layers are followed by two linear layers with 1000 hidden units. A single linear layer is attached in the case of $f_{EBR}$, and two linear layers for $\boldsymbol{\mu}$ and $\log \boldsymbol{\sigma}$, as in Eq.(\ref{eq:VAE_sampling}). The dimension of the embeddings, $\boldsymbol{\mu}$ and $\log \boldsymbol{\sigma}$ are all set to 256.


\subsection{Incremental Learning with Embedding Rehearsal}
\label{subsec:exp_ebr}

\begin{figure}[!ht]
    \centering
    \includegraphics[width=0.5\textwidth]{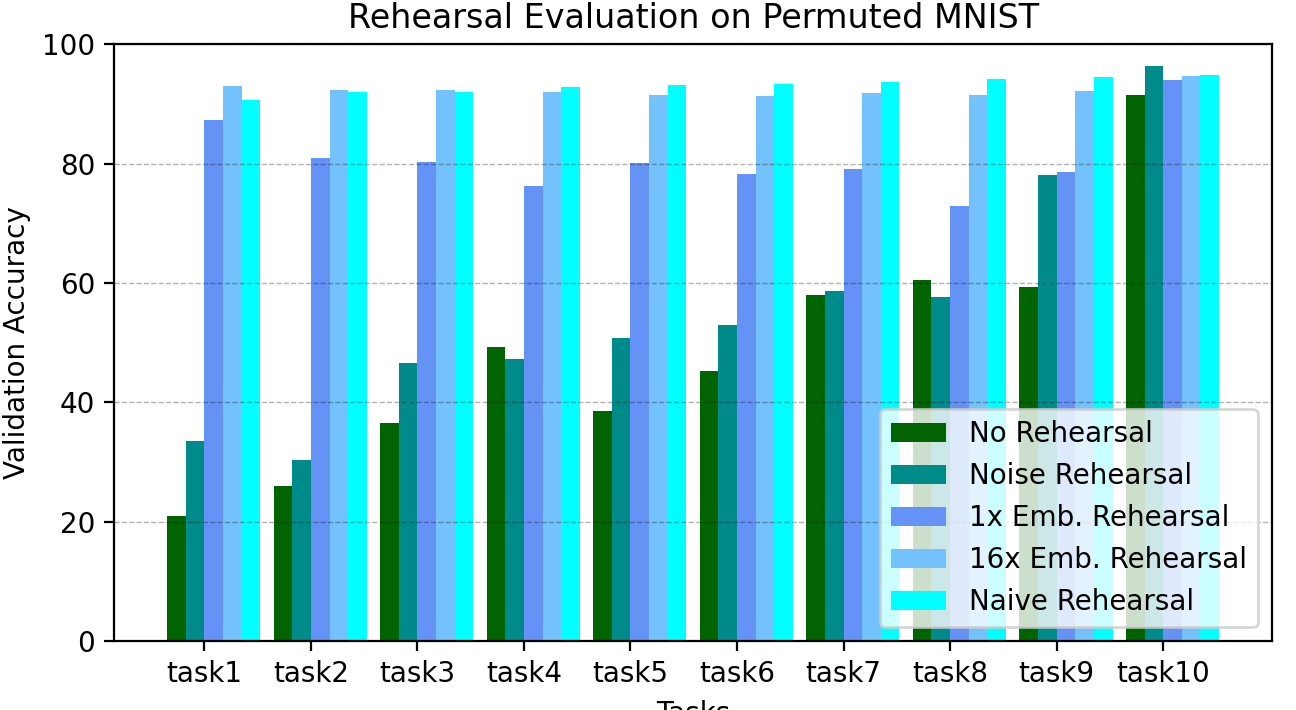}
    \caption{Validation accuracy on permuted MNIST based on rehearsal strategies.}
    \label{fig:permuted_MNIST}
\end{figure}

The ``\textit{Embedding-based Rehearsal}'' (EBR) is proven to be an efficient approach for Incremental Learning, when local user privacy constraints are not an issue\footnote{Unlike the Federated Learning scenario, where the local data in any form cannot be shared with the server.}. Herein, we investigate the performance of EBR against 3 other scenarios: \textit{No Rehearsal}, \textit{Noise Rehearsal with random projections} and \textit{Naive Rehearsal}.  
We create a new dataset based on 10 different random transformations of the MNIST dataset to simulate 10 different, non-overlapping tasks, as in Figure~\ref{fig:permuted_MNIST}. The transformations are different enough to cause Catastrophic Forgetting when the model is sequentially trained on these 10 tasks. The models are tested  on matched datasets after the training phase is concluded\footnote{The test set is transformed according to the matching  transformation per task.}.

The \textit{No Rehearsal} method is the baseline approach with no mitigation of  Catastrophic Forgetting in place. As expected, the model forgets the first tasks and the classification performance appears significantly lower, as in Fig.~\ref{fig:permuted_MNIST}. Similarly, the ``Noise Rehersal'' with random projections, i.e. using a randomly initialized $\boldsymbol{\theta}$ of the encoder $f_{EBR}(\ \cdot\ ;\boldsymbol{\theta})$, doesn't mitigate the problem of Catastrophic Forgetting and the performance degrades significantly when moving away from the latter trained tasks. Interestingly, the random projections are quite efficient when training/testing sets match -- the performance is equivalent to the \textit{No Rehearsal} method. In both cases, there is no significant mitigation of the Catastrophic Forgetting phenomenon, as expected~\cite{hsu2018re}.

On the other hand, the algorithms based on  {\it rehearsal} show a more consistent performance across all tasks. The \textit{Naive Rehearsal} (NR) is based on recycling a subset of training examples sampled from past tasks during training. Herein, the $\mathcal{R}_{T, Reh.}$ set consists of $10\%$ of the training samples seen per iteration $T$. 
Although, the NR performance is consistent across different tasks, Fig.~\ref{fig:permuted_MNIST}, the method poses privacy-related challenges. 
On the other hand, the EBR appears performing as well as NR without the need of storing any examples -- only the sampled embeddings are necessary for replaying. Two different EBR scenarios are investigated depending on the amount of data saved for rehearsal. In the ``\textit{EBR 1x}'' scenario, the number of the embedding vectors matches to the number of data samples used for the NR method. \textit{EBR 1x} experiment investigates the performance given the same number of rehearsing  samples as NR. In the case of ``\textit{EBR 16x}'' scenario, the same amount of memory is used (since the embeddings have a much smaller footprint than the raw samples). 
In both case, the degradation is due to Catastrophic Forgetting, however rehearsing with more data, i.e. \textit{EBR 16x} scenario, outperforms the \textit{EBR 1x} one.

The rehearsal approach, either \textit{NR} or \textit{EBR}, is an efficient strategy to mitigate Catastrophic Forgetting, in Fig.~\ref{fig:permuted_MNIST}. However, both  methods are unrealistic in FL scenarios due to the privacy concerns. Herein, we have proposed  replacing the deterministic encoder $f^*_{EBR}(\cdot)$ with the VEE one $f^*_{VEE}(\cdot)$. All benefits from \textit{EBR} are now combined with privacy preserving rehearsing, Section~\ref{sec:proposed}.

\subsection{Federated Incremental Learning with VER}
\label{subsec:filver}

\begin{figure}[htb]
    \centering
    \includegraphics[width=0.5\textwidth]{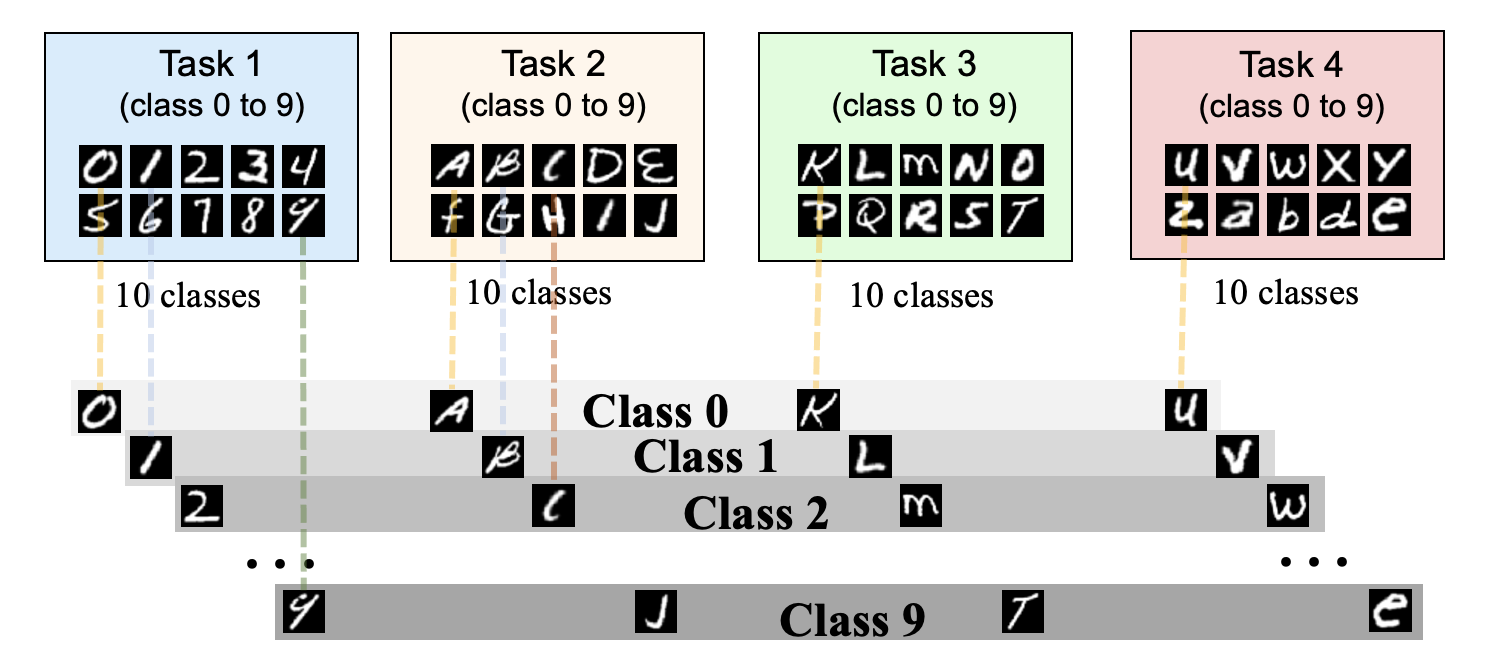}
    \caption{Incremental Task assignments with EMNIST dataset/classes.}
    \label{fig:4tasks}
\end{figure}

We use the EMNIST dataset~\cite{cohen2017emnist} to simulate a federated  ``Incremental Domain Learning'' (IDL) scenario. The EMNIST dataset is a set of handwritten characters and digits captured and converted to $28\times28$ pixel images, matching the well known MNIST dataset. Among the many splits of EMNIST, we use the ``EMNIST Balanced'', which contains 131,600 images with 47 balanced classes. We create four different, disjoint tasks, labeled ``\textit{Task 1}'', ``\textit{Task 2}'', ``\textit{Task 3}'' and ``\textit{Task 4}'', where the first 40 classes of the ``EMNIST Balanced'' split are uniformly distributed, as in Fig.~\ref{fig:4tasks}. 
For the IDL setup, we assume that any model or local FL client cannot have access to the previous tasks, e.g., if the model is training and predicting ``Task 4'', the model cannot have access to data from ``Tasks 1-3''. Each task matches the original MNIST classification setup, where the model classifies 10 classes\footnote{This task is designed to investigate  model drifting, not focused on providing state-of-the-art handwritten character recognition performance.}. Each client contains balanced training examples per label, sequentially per task. The model is trained with 50 rounds per task, before switching to the next task -- data from previous tasks are then discarded. We pretrain the VEE on the first task of the process, i.e. no additional held-out data is used for pre-training the model. Henceforth, the encoder parameters $f^*_{VEE}(\cdot)$ remain frozen in all subsequent training iterations.

\subsubsection{FIL-VER Experiments}
\label{subsubsec:FIL-VER}

First, we investigate the performance of \textit{FIL-VER} when all clients are always enrolled in the system, i.e. ``\textit{Scenario 1}'' in DCE, Section~\ref{sec:dynamic_enroll}. The 4 systems evaluated are ``\textit{Vanilla FL without rehearsal}'', ``\textit{VER w/o SST}'', ``\textit{VER Stats}'', and finally, ``\textit{VER  Sampled}''. The benchmark performance is based on the ''\textit{Offline Val. Acc.}'' system where the model is trained offline on data from all 4 tasks. The method \textit{VER Stats} indicates the mean and variance values are sent to the server to generate embeddings for rehearsal. In the case of \textit{VER Sampled} the clients send sampled embedding vectors that are generated from the statistics retained on the clients -- without ever revealing these statistics to the server. As such, the method \textit{VER Sampled} is the method proposed to prevent sharing deterministically the clients' data  either by sending either the raw embeddings or the mean/variance values. 

\begin{figure*}[htb]
    \begin{subfigure}[t]{0.5\textwidth}
      \includegraphics[width=\linewidth]{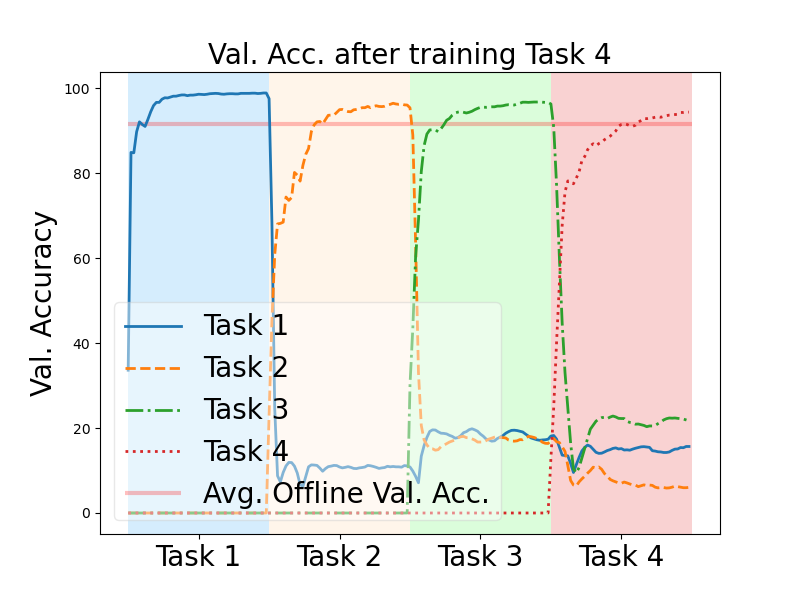}
      \caption{Vanilla Fed. Learning}
      \label{fig:scenario1_vanilla_fed}
    \end{subfigure}
    \hfill
    \begin{subfigure}[t]{0.5\textwidth}
      \includegraphics[width=\linewidth]{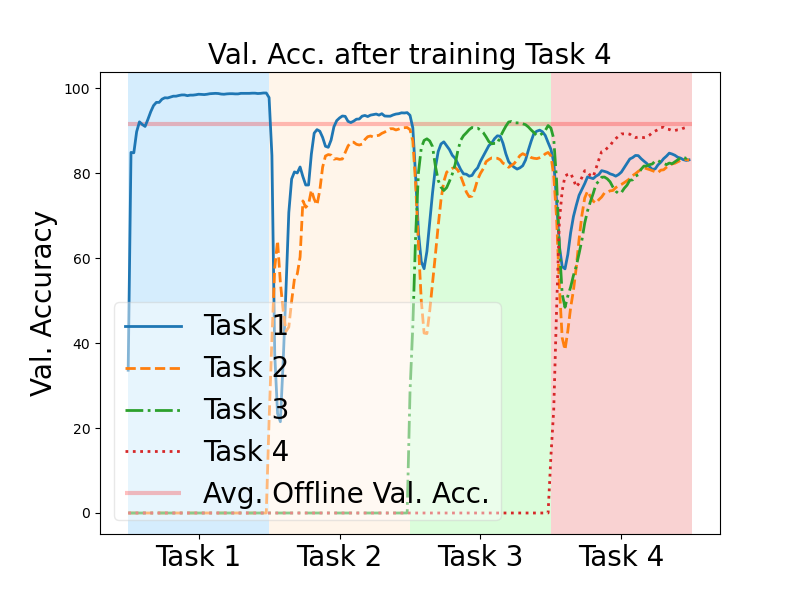}
      \caption{Client-side VER Without SST}
      \label{fig:scenario1_without_sst}
    \end{subfigure}
    \vskip\baselineskip
    \begin{subfigure}[t]{0.5\textwidth}
      \includegraphics[width=\linewidth]{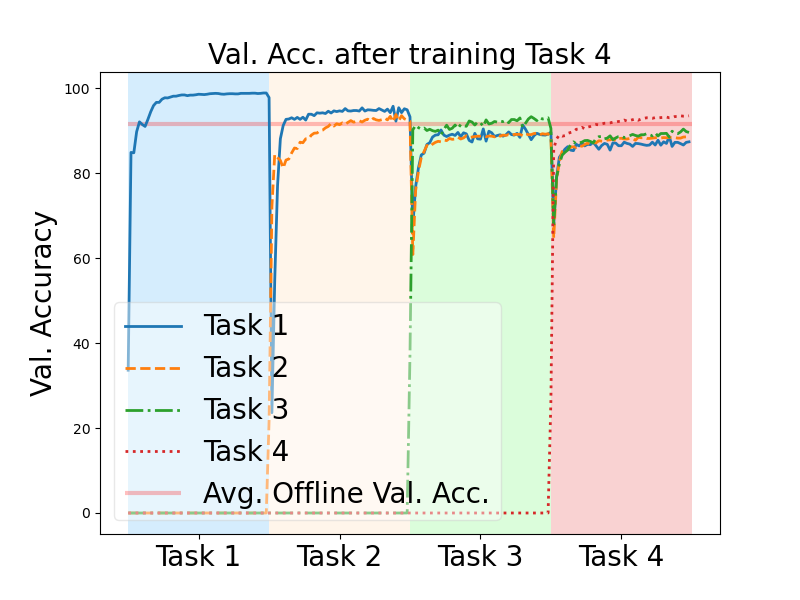}
      \caption{VER Sharing Stats}
      \label{fig:scenario1_mgn_stats}
    \end{subfigure}
    \hfill
    \begin{subfigure}[t]{0.5\textwidth}
      \includegraphics[width=\linewidth]{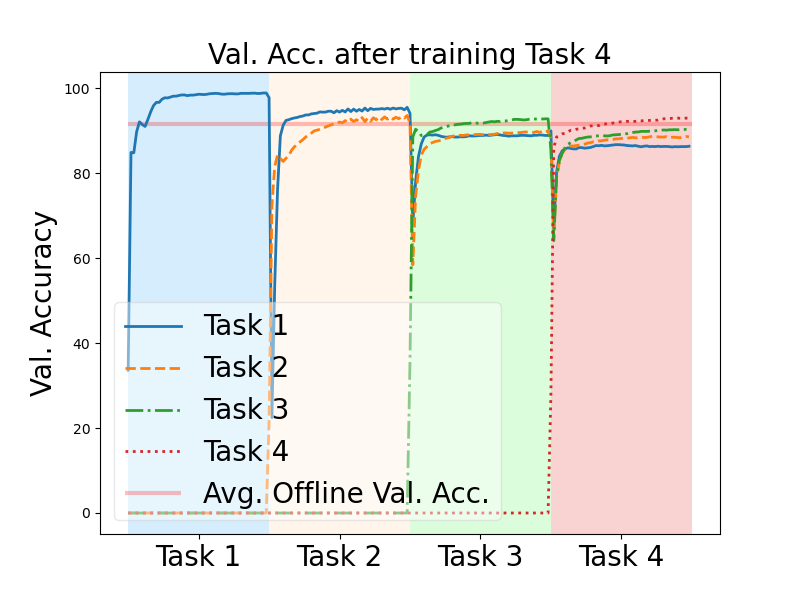}
      \caption{VER Sharing Samples}
      \label{fig:scenario1_mgn_sampled}
    \end{subfigure}
\caption{Validation accuracy of VER: Scenario 1. The model for each task is trained for 50 iterations.}
\label{fig:training_plot}
\end{figure*}

As expected, Section~\ref{subsec:exp_ebr}, the performance of the \textit{Vanilla Fed.} system is the worst and the system quickly forgets all the previous tasks, Fig.~\ref{fig:scenario1_vanilla_fed}. The degradation is almost instantaneous once the model starts learning the new tasks. 

In the case of \textit{VER w/o SST} the system appears more prone to Catastrophic Forgetting, Fig.~\ref{fig:scenario1_without_sst}. As shown, the non-deterministic VEE embeddings do not avert the classifier from learning. However, the lack of SST is detrimental to the final model performance. The overall performance, although  lower than that of the benchmark system, is still able to compensate for the IDL due to the rehearsal step taking place  on the clients (but not on the server). 

On the other hand,  the proposed SST step, in both cases of \textit{VER Stats} and \textit{VER Sampled}, enables a model performance close to the offline system, Figs.~\ref{fig:scenario1_mgn_stats} and~\ref{fig:scenario1_mgn_sampled}, without much fluctuation in early tasks. Also, the  \textit{VER Sampled} method achieves good validation accuracy without sharing  the actual statistics (containing the private information). 
%

Concluding, the \textit{VER Stats} method might not be ideal based on the privacy constraints,  since these stats can leak private information. On the other hand, rehearsing on sampled VEE embeddings doesn't pose such problem, as discussed in Section~\ref{subsec:SST}.  Despite the fact that the proposed approach doesn't follow rigorous formulation for Differential Privacy~\cite{DMNS06}, the methodology has potential to branch out to other applications where privacy-protected representations is required for FL.

\subsubsection{Dynamic Client Enrollment Scenarios}
\label{subsubsec:DES}

\begin{figure*}[htb]
    \begin{subfigure}[t]{0.5\textwidth}
      \includegraphics[width=\linewidth]{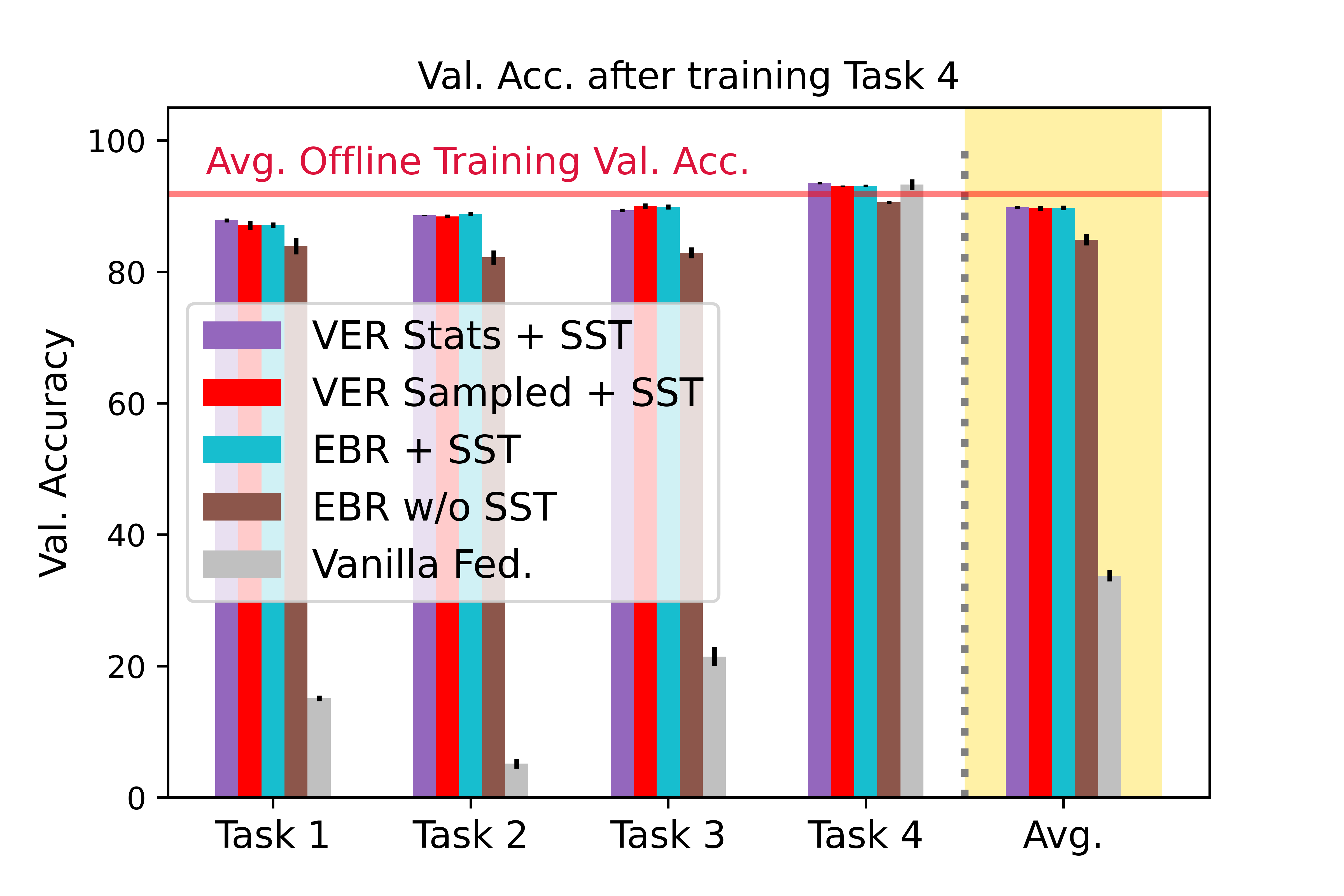}
      \caption{Validation accuracy after task \#4 in Scenario 1.}
      \label{fig:scenario1_val_accs}
    \end{subfigure}
    \hfill
    \begin{subfigure}[t]{0.5\textwidth}
      \includegraphics[width=\linewidth]{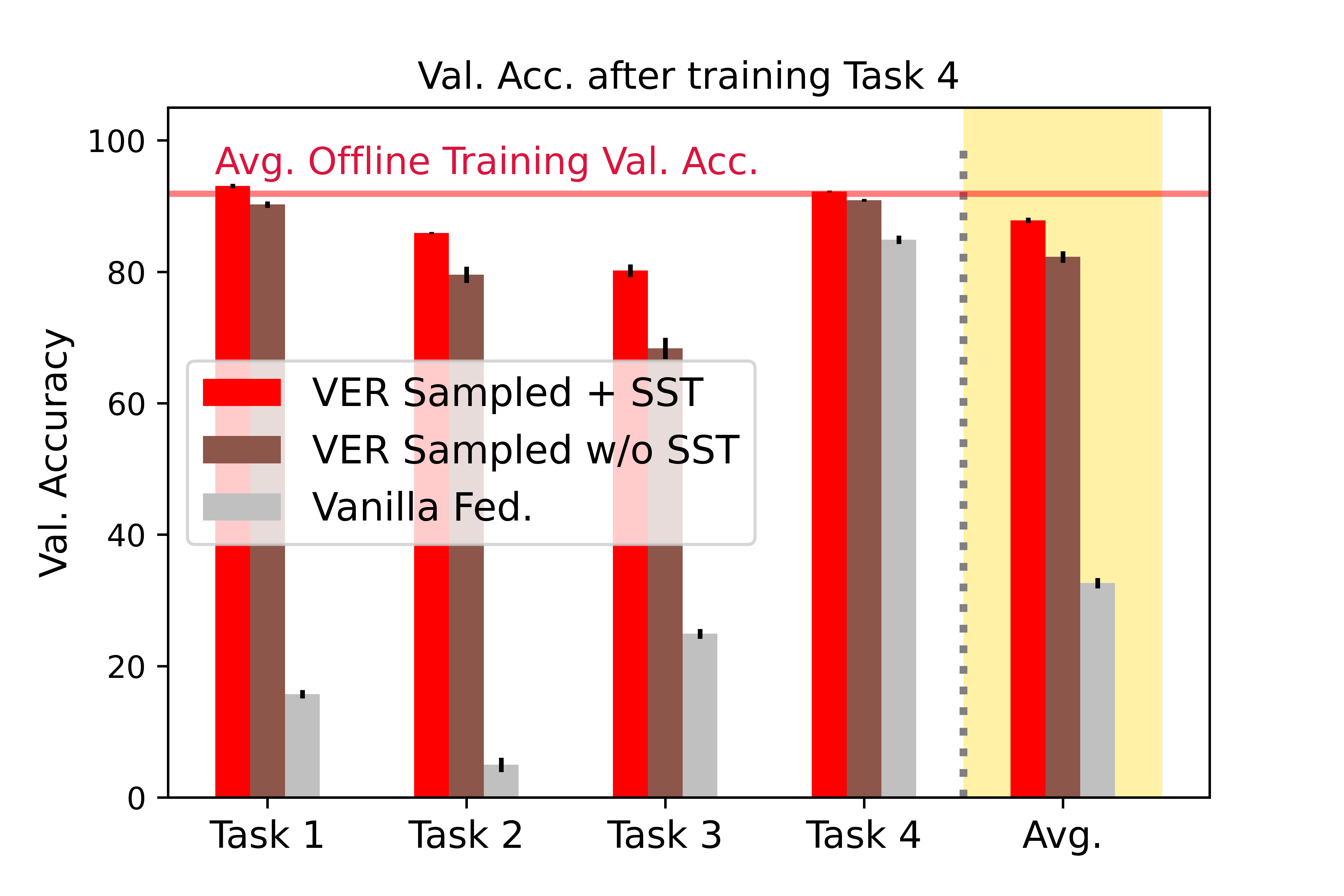}
      \caption{Validation accuracy after task \#4 in Scenario 2.}
      \label{fig:scenario2_val_accs}
    \end{subfigure}
    \vskip\baselineskip
    \begin{subfigure}[t]{0.5\textwidth}
      \includegraphics[width=\linewidth]{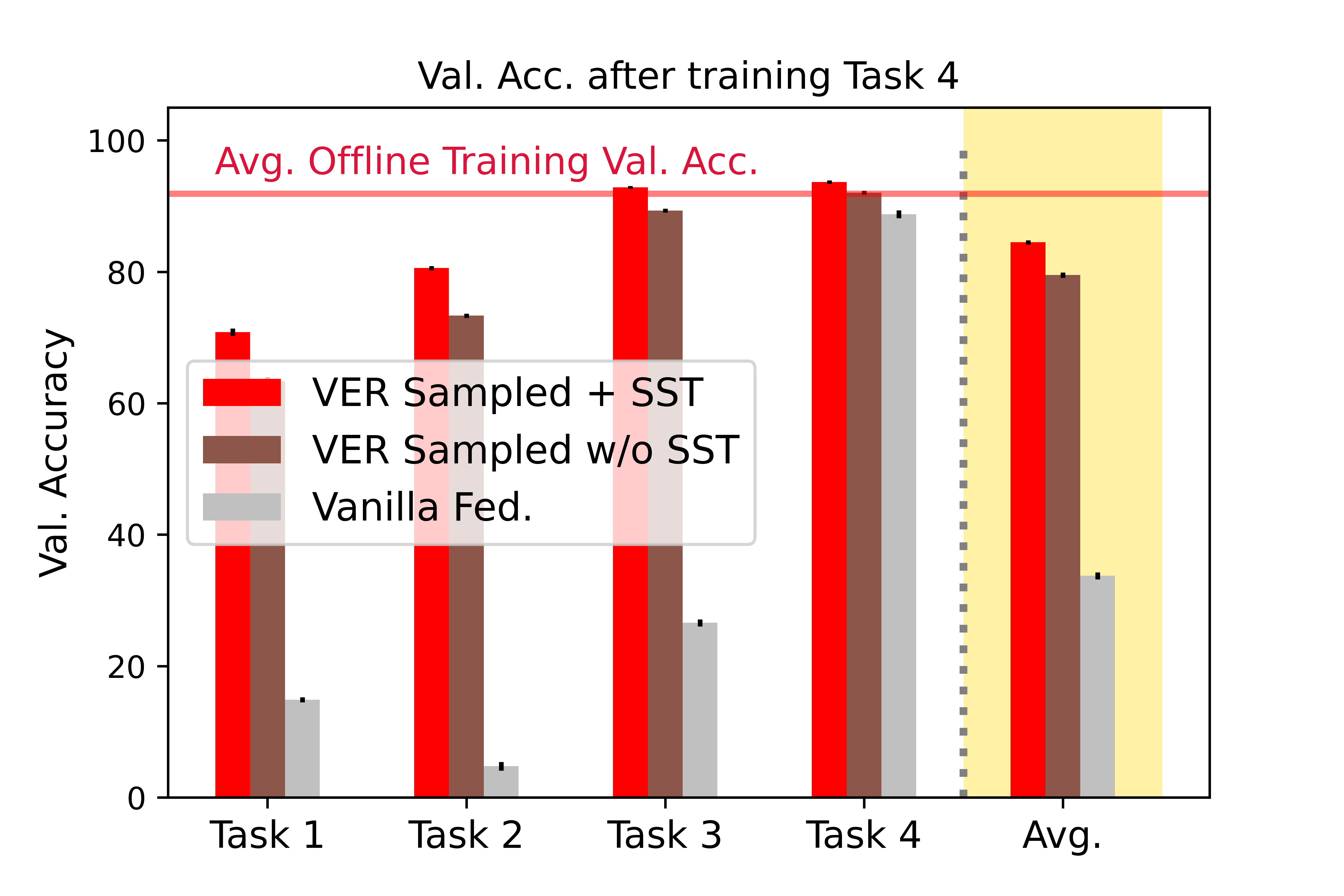}
      \caption{Validation accuracy after task \#4 in Scenario 3.}
      \label{fig:scenario3_val_accs}
    \end{subfigure}
    \hfill
    \begin{subfigure}[t]{0.5\textwidth}
      \includegraphics[width=\linewidth]{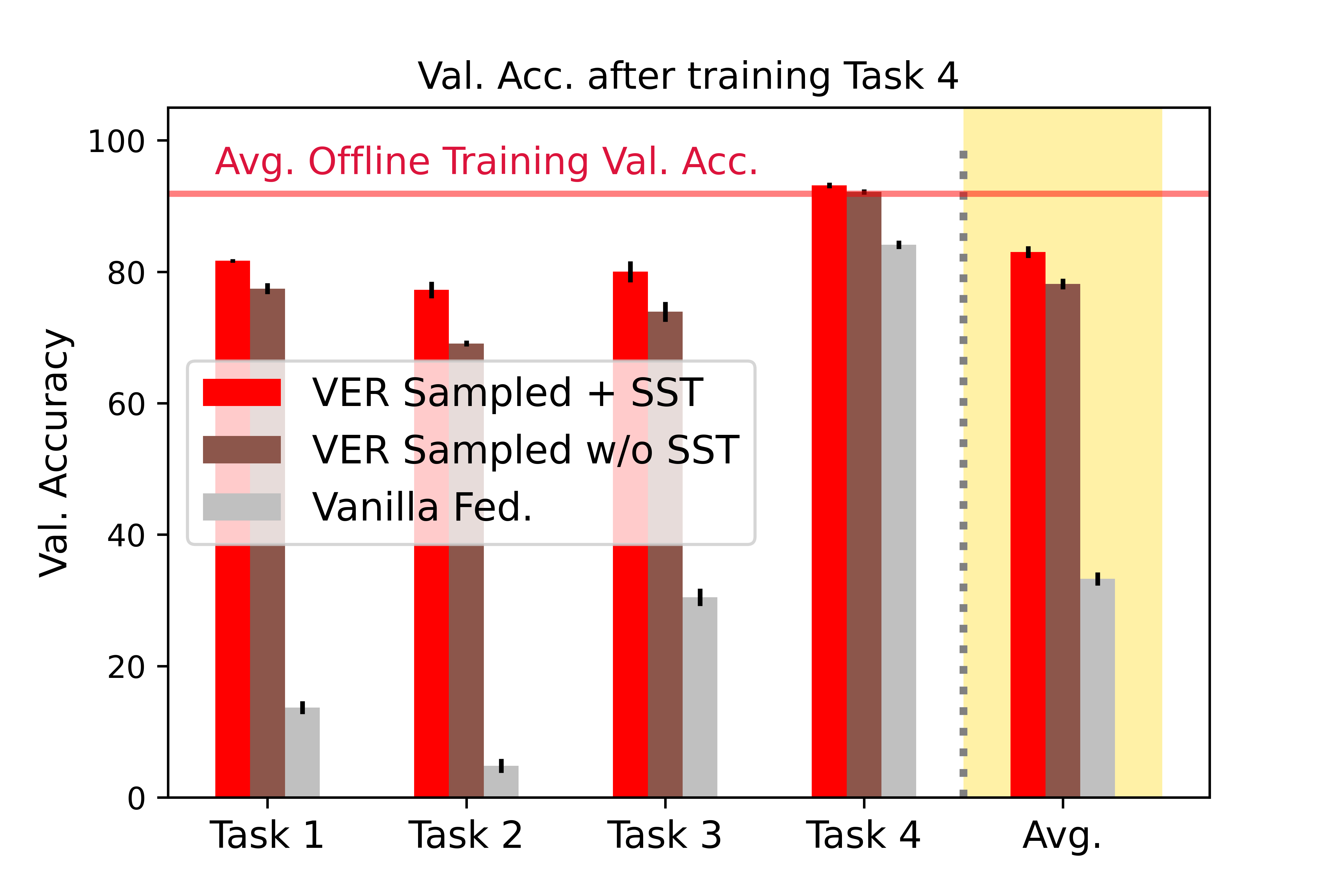}
      \caption{Validation accuracy after task \#4 in Scenario 4.}
      \label{fig:scenario4_val_accs}
    \end{subfigure}
\caption{Accuracy after sequentially training on Task-1 to Task-4: Comparison of Performance of Incremental Variational Rehearsal with/without SST against the baseline FL system.}
\label{fig:scenario_val_accs}
\end{figure*}

Lastly, the FIL-VER framework is investigated as part of the DCE scenarios, Sections~\ref{sec:dynamic_enroll}, where the clients can dynamically enroll in or drop out of the system. Although such analysis is closer to real-life cases, where the number of clients providing the data can vary significantly, it is vastly ignored by most FL systems. Adding or losing a significant number of clients can have impact on the overall model training process. Contrary to the conventional setups for federated learning systems, the experimental results from \textit{Scenarios 2-4} show what are the effects of the varying number of client enrollments in the process. The proposed SST approach addresses such scenarios by rehearsing on the embeddings collected from the clients even when these clients no longer participate. However, this SST approach shouldn't be performed without privacy-enabled methods such as the proposed VEE-based representations. As shown in Figure~\ref{fig:scenario_val_accs}, both SST and VER approaches are essential components for various DCE scenarios. 

\paragraph{Scenario 1: Fully Enrolled}
The performance on the validation sets for ``Tasks 1-4''  is shown in Fig. \ref{fig:scenario1_val_accs}, when all clients are enrolled from the beginning, i.e. \textit{Scenario 1}.  The \textit{Avg.} accuracy is estimated over all four tasks. Different setups are compared, with the proposed FIL-VER method outperforming all baselines. As before, \textit{Vanilla Fed.} is the typical FL setup without any CL method employed. It shows a relatively poor performance dropping below 10\% of accuracy in early tasks, e.g. ``Task 2'', since there is no rehearsal. 
The \textit{EBR} method shares the original embedding vectors with the server. The \textit{w/o SST} identifier is used when there is no SST, and training relies solely on client-side training and rehearsing. Despite that, the \textit{EBR w/o SST}  still shows some robustness. However, the \textit{Avg.} validation accuracy remains lower than the \textit{EBR+SST} method mainly because the clients use their local embeddings while not taking advantage of data points other clients have.

\paragraph{Scenario 2: Decreasing Client base}
For the case of \textit{Scenarios 2, 3 and 4} we only show \textit{Vanilla Fed.} \textit{VER + SST} and \textit{VER w/o SST} for brevity. The number of enrolled clients decrease with time in \textit{Scenario 2}. Since, most of the active clients participated in ``Task 1'' (it is the earliest task) providing rehearsal data for SST, ``Task 1'' accuracy appears relatively elevated. On the contrary, training data points for ``Task 2-4'' are  limited (compared with ``Task 1'') since some of the clients have dropped out. In addition, the validation accuracy drops significantly when there is no SST, as shown in Section~\ref{subsubsec:FIL-VER}. While overall performance is lower than \textit{Scenario 1}, we can see that the performance drop in \textit{VER Sampled w/o SST} is even more significant than \textit{Scenario 1} where all clients can provide training data for all tasks. 

\paragraph{Scenario 3: Increasing Client base}
An increasing number of enrolled clients per each task is investigated in \textit{Scenario 3}, opposite to \textit{Scenario 2}. The system has incrementally more training data for latter tasks, e.g., ``Task 3'' has more training data points that ``Task 1''. Since, the model is trained on more data points from ``Task 4'' than from ``Task 1-3'', the validation accuracy increases with the task index increasing, Fig.~\ref{fig:scenario3_val_accs}. Also, the trend of repeated training data-points is  accelerated with SST, where ``Task 1'' shows higher validation accuracy. All in all, the \textit{Avg.} accuracy of the overall trial is similar to that of \textit{Scenario 2} but worse than that of \textit{Scenario 1}.

\paragraph{Scenario 4: Scattered Client base}
\textit{Scenario 4} is the most challenging case where the clients participate only for a short time window -- herein the FL system has a single client per task. Consequently, the overall validation accuracy is relatively lower that \textit{Scenario 1} showing even larger performance degradation for earlier tasks, ``Task 1'' or ``Task 2''. However, SST can compensate the performance even when the model is trained on less data per task. The average accuracy of \textit{Scenario 4} is no different than of \textit{Scenario 2}. This indicates that the proposed VER method has the capability of mitigating the effect of sparse clients' data by sharing the clients' representations with the server.

\section{Conclusions}
\label{sec:conclusions}
In this paper, we propose a novel FL method to mitigate the problem of Catastrophic Forgetting. The proposed \textit{Federated Incremental Learning with Variational Embeddings Rehearsal} framework has two notable advantages: first, the proposed method significantly suppresses the model forgetting the previous knowledge, especially in a federated learning setup. Such approach is enhanced by employing \textit{Server-Side Training}, where client-based sampled data-points are used for rehearsing on the server side. Second, our proposed FIL-VER framework provides privacy guaranties with the estimated embeddings are decoupled from the raw input data.
Embedding-based rehearsal, i.e. the \textit{EBR}, \textit{VER Stats} and \textit{VER Sampled} methods show small differences in overall performance, as in Figs. \ref{fig:scenario1_val_accs} and~\ref{fig:training_plot}. However, the VEE-based embeddings ensure privacy-protected FL system while addressing the Catastrophic Forgetting.

Also, a novel \textit{Dynamic Client Enrollment} process is introduced, investigating how an FL system performance is affected by the number of participating clients. The proposed \textit{FIL-VER} framework can still mitigate the Catastrophic Forgetting, while different DCE scenarios are evaluated. 

Future work includes a scaled experimental setup of the proposed method, for example, text sentiment analysis, face recognition and speaker recognition can be good candidates since these applications exhibit potential risk of infringing clients' privacy. In addition to these experiments, a different research direction can be followed to investigate  task-invariant embedding extraction, such as the IRM framework. We haven't investigated so far  the impact of the volume of data sent from clients to the server. Finally, there is still a lack of theoretical warranties like in the Differential Privacy case despite the proposed system has potential to be applied in many different FL system where privacy and online learning is a crucial aspect.

\nocite{langley00}

\bibliography{main_icml}

\begin{thebibliography}{25}
\providecommand{\natexlab}[1]{#1}
\providecommand{\url}[1]{\texttt{#1}}
\expandafter\ifx\csname urlstyle\endcsname\relax
  \providecommand{\doi}[1]{doi: #1}\else
  \providecommand{\doi}{doi: \begingroup \urlstyle{rm}\Url}\fi

\bibitem[Arjovsky et~al.(2020)Arjovsky, Bottou, Gulrajani, and
  Lopez-Paz]{ABGL20}
Arjovsky, M., Bottou, L., Gulrajani, I., and Lopez-Paz, D.
\newblock Invariant risk minimization.
\newblock \emph{arXiv preprint arXiv:1907.02893}, 2020.

\bibitem[Casado et~al.(2021)Casado, Lema, Criado, Iglesias~Rodriguez, Regueiro,
  and Barro]{Casado+21}
Casado, F., Lema, D., Criado, M., Iglesias~Rodriguez, R., Regueiro, C., and
  Barro, S.
\newblock Concept drift detection and adaptation for federated and continual
  learning.
\newblock \emph{Multimedia Tools and Applications}, 07 2021.

\bibitem[Cohen et~al.(2017)Cohen, Afshar, Tapson, and
  Van~Schaik]{cohen2017emnist}
Cohen, G., Afshar, S., Tapson, J., and Van~Schaik, A.
\newblock Emnist: Extending mnist to handwritten letters.
\newblock In \emph{2017 International Joint Conference on Neural Networks
  (IJCNN)}, pp.\  2921--2926. IEEE, 2017.

\bibitem[Cole et~al.(2017)Cole, Belanger, Krishnan, Sarna, Mosseri, and
  Freeman]{cole2017synthesizing}
Cole, F., Belanger, D., Krishnan, D., Sarna, A., Mosseri, I., and Freeman,
  W.~T.
\newblock Synthesizing normalized faces from facial identity features.
\newblock In \emph{Proceedings of the IEEE conference on computer vision and
  pattern recognition}, pp.\  3703--3712, 2017.

\bibitem[Delange et~al.(2021)Delange, Aljundi, Masana, Parisot, Jia, Leonardis,
  Slabaugh, and Tuytelaars]{Delange+21}
Delange, M., Aljundi, R., Masana, M., Parisot, S., Jia, X., Leonardis, A.,
  Slabaugh, G., and Tuytelaars, T.
\newblock {A continual learning survey: Defying forgetting in classification
  tasks}.
\newblock \emph{IEEE Trans. on Pattern Analysis and Machine Intell.}, 2021.

\bibitem[Dwork et~al.(2006)Dwork, McSherry, Nissim, and Smith]{DMNS06}
Dwork, C., McSherry, F., Nissim, K., and Smith, A.
\newblock Calibrating noise to sensitivity in private data analysis.
\newblock In \emph{Theory of Cryptography}, pp.\  265--284. Springer Berlin
  Heidelberg, 2006.

\bibitem[Francis et~al.(2021)Francis, Tenison, and Rish]{FrTeRi21}
Francis, S., Tenison, I., and Rish, I.
\newblock Towards causal federated learning for enhanced robustness and
  privacy.
\newblock \emph{arXiv preprint arXiv:2104.06557}, 2021.

\bibitem[French(1999)]{french1999catastrophic}
French, R.~M.
\newblock Catastrophic forgetting in connectionist networks.
\newblock \emph{Trends in cognitive sciences}, 3\penalty0 (4):\penalty0
  128--135, 1999.

\bibitem[Hsu et~al.(2018)Hsu, Liu, Ramasamy, and Kira]{hsu2018re}
Hsu, Y.-C., Liu, Y.-C., Ramasamy, A., and Kira, Z.
\newblock Re-evaluating continual learning scenarios: A categorization and case
  for strong baselines.
\newblock \emph{arXiv preprint arXiv:1810.12488}, 2018.

\bibitem[Javed \& White(2019)Javed and White]{JaWh19}
Javed, K. and White, M.
\newblock Continual learning with deep generative replay.
\newblock In \emph{Proc. NeurIPS}, 2019.

\bibitem[Jiang et~al.(2017)Jiang, Zheng, Tan, Tang, and Zhou]{JIANG+17}
Jiang, Z., Zheng, Y., Tan, H., Tang, B., and Zhou, H.
\newblock Variational deep embedding: An unsupervised and generative approach
  to clustering.
\newblock In \emph{Proc. of IJCAI'17}. AAAI Press, 2017.

\bibitem[Kairouz \& al(2021)Kairouz and al]{Kairouz+21}
Kairouz, P. and al.
\newblock Advances and open problems in federated learning.
\newblock \emph{Foundations and Trends in Machine Learning}, 14\penalty0
  (1–2):\penalty0 1--210, 2021.

\bibitem[Kingma \& Welling(2013)Kingma and Welling]{kingma2013auto}
Kingma, D.~P. and Welling, M.
\newblock Auto-encoding variational bayes.
\newblock \emph{arXiv preprint arXiv:1312.6114}, 2013.

\bibitem[Kingma \& Welling(2019)Kingma and Welling]{kingma2019introduction}
Kingma, D.~P. and Welling, M.
\newblock An introduction to variational autoencoders.
\newblock \emph{arXiv preprint arXiv:1906.02691}, 2019.

\bibitem[Konecny et~al.(2015)Konecny, McMahan, and Ramage]{KMR15}
Konecny, J., McMahan, B.~H., and Ramage, D.
\newblock {Federated Optimization: Distributed Optimization Beyond the
  Datacenter}.
\newblock \emph{arXiv preprint arXiv:1511.03575v1}, 2015.

\bibitem[Langley(2000)]{langley00}
Langley, P.
\newblock Crafting papers on machine learning.
\newblock In Langley, P. (ed.), \emph{Proceedings of the 17th International
  Conference on Machine Learning (ICML 2000)}, pp.\  1207--1216, Stanford, CA,
  2000. Morgan Kaufmann.

\bibitem[Li et~al.(2020)Li, Wen, Wu, Hu, Wang, and He]{Li+20}
Li, Q., Wen, Z., Wu, Z., Hu, S., Wang, N., and He, B.
\newblock {A Survey on Federated Learning Systems: Vision, Hype and Reality for
  Data Privacy and Protection}.
\newblock \emph{arXiv preprint arXiv:1907.09693v4}, 2020.

\bibitem[McMahan et~al.(2017)McMahan, Moore, Ramage, Hampson, and
  Arcas]{McMahan+17}
McMahan, H.~B., Moore, E., Ramage, D., Hampson, S., and Arcas, B. A.~y.
\newblock {Communication-efficient Learning of Deep Networks from Decentralized
  Data}.
\newblock In \emph{Proc. International Conference on Artificial Intelligence
  and Statistics}, pp.\  1273--–1282, 2017.

\bibitem[Pomponi et~al.(2021)Pomponi, Scardapane, and Uncini]{PSU21}
Pomponi, J., Scardapane, S., and Uncini, A.
\newblock Pseudo-rehearsal for continual learning with normalizing flows.
\newblock \emph{arXiv preprint arXiv:2007.02443}, 2021.

\bibitem[Rebuffi et~al.(2017)Rebuffi, Kolesnikov, Sperl, and Lampert]{RKSL17}
Rebuffi, S.~A., Kolesnikov, A., Sperl, G., and Lampert, C.~H.
\newblock icarl: Incremental classifier and representation learning.
\newblock In \emph{Proc. CVPR}, 2017.

\bibitem[Rezende et~al.(2014)Rezende, Mohamed, and
  Wierstra]{rezende2014stochastic}
Rezende, D.~J., Mohamed, S., and Wierstra, D.
\newblock Stochastic backpropagation and approximate inference in deep
  generative models.
\newblock In \emph{International conference on machine learning}, pp.\
  1278--1286. PMLR, 2014.

\bibitem[Shin et~al.(2017)Shin, Lee, Kim, and Kim]{SLKK17}
Shin, H., Lee, J.~K., Kim, J., and Kim, J.
\newblock Continual learning with deep generative replay.
\newblock In \emph{Proc. NIPS}, 2017.

\bibitem[Wang \& al(2021)Wang and al]{wang+21}
Wang, J. and al.
\newblock A field guide to federated optimization.
\newblock \emph{arXiv preprint arXiv:2107.06917}, 2021.

\bibitem[Yoon et~al.(2020)Yoon, Jeong, Lee, Yang, and Hwang]{yoon2020federated}
Yoon, J., Jeong, W., Lee, G., Yang, E., and Hwang, S.~J.
\newblock Federated continual learning with adaptive parameter communication.
\newblock \emph{arXiv preprint arXiv:2003.03196}, 2020.

\bibitem[Zhmoginov \& Sandler(2016)Zhmoginov and
  Sandler]{zhmoginov2016inverting}
Zhmoginov, A. and Sandler, M.
\newblock Inverting face embeddings with convolutional neural networks.
\newblock \emph{arXiv preprint arXiv:1606.04189}, 2016.

\end{thebibliography}
\bibliographystyle{icml2020}


\end{document}